\documentclass[letterpaper]{article} 
\usepackage{aaai2026}  
\usepackage{times}  
\usepackage{helvet}  
\usepackage{courier}  
\usepackage[hyphens]{url}  
\usepackage{graphicx} 
\usepackage{natbib}  
\usepackage{caption} 
\usepackage{algorithm}
\usepackage{algorithmic}

\usepackage{subcaption}
\usepackage{xcolor}
\usepackage{amsmath,amssymb}
\usepackage{booktabs}
\usepackage{multirow}

\usepackage{newfloat}
\usepackage{listings}
\DeclareCaptionStyle{ruled}{labelfont=normalfont,labelsep=colon,strut=off} 
\floatstyle{ruled}
\newfloat{listing}{tb}{lst}{}
\floatname{listing}{Listing}
\title{GuidNoise: Single-Pair Guided Diffusion for Generalized Noise Synthesis}
\author {
    Changjin Kim\textsuperscript{\rm 1},
    HyeokJun Lee\textsuperscript{\rm 2},
    YoungJoon Yoo\textsuperscript{\rm 1, \rm 2}
}
\affiliations {
    \textsuperscript{\rm 1}SNUAILAB\\
    \textsuperscript{\rm 2}Dept. of Artificial Intelligence, Chung-Ang University\\
    cjkim@snuailab.ai, \{gurwns8926, yjyoo3312\}@cau.ac.kr
}

\begin{document}

\maketitle

\begin{abstract}
\label{abstract}
Recent image denoising methods have leveraged generative modeling for real noise synthesis to address the costly acquisition of real-world noisy data. However, these generative models typically require camera metadata and extensive target-specific noisy-clean image pairs, often showing limited generalization between settings.
In this paper, to mitigate the prerequisites, we propose a Single-Pair Guided Diffusion for generalized noise synthesis(\textbf{GuidNoise}), which uses a \textbf{single noisy/clean pair} as the guidance, often easily obtained by itself within a training set. 
To train GuidNoise, which generates synthetic noisy images from the guidance, we introduce a guidance-aware affine feature modification (GAFM) and a noise-aware refine loss to leverage the inherent potential of diffusion models.
This loss function refines the diffusion model's backward process, making the model more adept at generating realistic noise distributions. 
The GuidNoise synthesizes high-quality noisy images under diverse noise environments without additional metadata during both training and inference. Additionally, GuidNoise enables the efficient generation of noisy-clean image pairs at inference time, making synthetic noise readily applicable for augmenting training data. This self-augmentation significantly improves denoising performance, especially in practical scenarios with lightweight models and limited training data. The code is available at https://github.com/chjinny/GuidNoise.
\end{abstract}
    
\section{Introduction}
\label{sec:introduction}
Image denoising is a crucial task in computer vision, enhancing image quality and facilitating higher-level vision tasks. 
The main challenge lies in effectively removing noise while preserving image details, especially in real-world conditions where noise patterns are influenced by the complexities of modern camera systems such as an image signal processing (ISP) pipeline.

\begin{figure}
    \centering
    \includegraphics[trim=1 1 1 1, clip, width=0.37\textwidth]{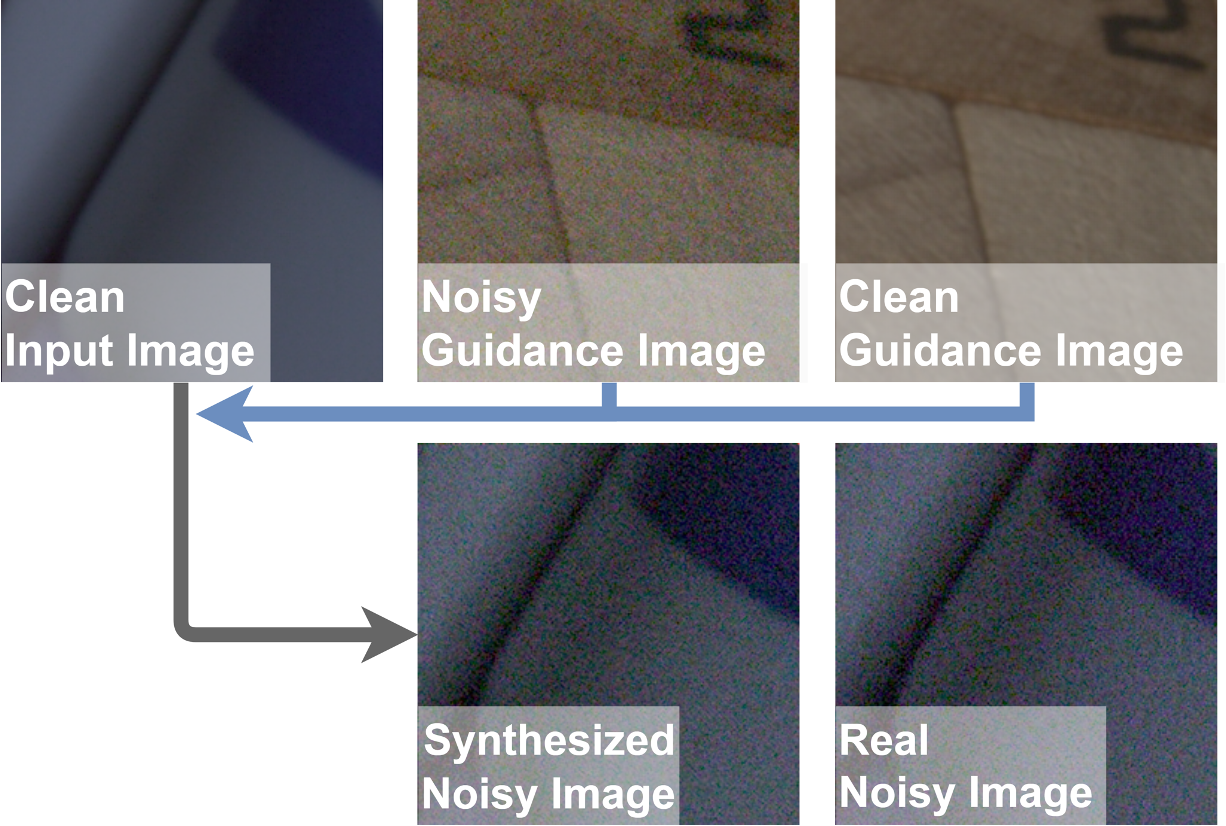}
    \caption{
    \textbf{Illustration of GuidNoise.} Given an input clean image and a noisy-clean guidance image pair, GuidNoise can generate a synthesized noisy image that mimics the real noisy image by capturing the noise distribution of the guidance image. Since the given images can be easily obtained from diverse environments, pseudo-real noisy images can be synthesized arbitrarily.
    }
    \label{fig:enter-label}
\end{figure}

Deep neural networks, particularly convolutional neural networks (CNN), have significantly advanced image denoising~\cite{DnCNN:zhang2017beyond, RIDNet:anwar2019real}. Initially, image-denoising research relied on basic synthetic noise models, primarily trained on typical noise distributions such as Additive White Gaussian Noise (AWGN). These models perform well in controlled environments, but often struggle to generalize to noisy real-world images with complex noise patterns. 
To address this limitation, datasets consisting of real-world noisy images have been proposed, such as SIDD~\cite{SIDD:SIDD_2018_CVPR} and PolyU~\cite{PolyU:xu2018real}. However, creating these datasets is costly because it requires an extensive collection of real noisy images and the corresponding acquisition of clean images paired with the target noisy image.

To reduce the costs associated with preparing noisy image data sets in real-world situations, various generative models have been proposed to synthesize noise by learning from real-world data.
Specifically, Generative Adversarial Networks (GAN)~\cite{goodfellow2020generative}, Normalizing Flow (NF)~\cite{papamakarios2021normalizing}, and diffusion~\cite{ho2020denoising} models have shown promising results~\cite{NoiseFlow:abdelhamed2019noise, DANet_AKLD_KLD_PGap:yue2020dual, CycleISP:zamir2020cycleisp, C2N:jang2021c2n, PNGAN:cai2021learning, Flow-sRGB:kousha2022modeling, NeCA:fu2023srgb, NAFlow:kim2023srgb,wu2025realistic}.
The recent noise synthesis models utilize various techniques using camera metadata or a number of noisy/clean image pairs of the target scene to achieve higher noise modeling performance.

Specifically, a line of approaches \cite{NoiseFlow:abdelhamed2019noise, Flow-sRGB:kousha2022modeling, NeCA:fu2023srgb} incorporates camera-specific information to generate noise that closely mimics the characteristics of particular imaging systems, while the others \cite{DANet_AKLD_KLD_PGap:yue2020dual, NAFlow:kim2023srgb} use target noisy images paired with input clean images to capture the intricate patterns of noise distributions.
These methods have demonstrated superior performance in generating high-fidelity noise, effectively capturing the complexities of actual camera outputs.
However, while these methods perform successfully under the assumption of equivalent environments between training and testing, they often struggle to generalize across diverse conditions, including different devices, ISO settings, shutter speeds, and scene variations. 
Their reliance on specific metadata or a large number of paired images can limit their adaptation capability to broader datasets or scenarios where such detailed information is unavailable. 

To address the limitations, we propose a single-pair guided diffusion model for noise synthesis named \textbf{GuidNoise}. As illustrated in Figure~\ref{fig:enter-label}, GuidNoise trains the diffusion model to generate noise that mimics one of the noise distributions contained in a guidance image pair. At inference time, it requires only a single noisy-clean image pair from the target domain and shows high adaptability across diverse datasets.
We note that our method only requires a single guidance pair for the noise synthesis and significantly relaxes the constraint of acquiring clean/noisy image pairs or getting camera metadata.
At the inference phase in GuidNoise, a generator can inject a noise distribution contained in a guidance image pair from target domain into every given real clean image, without actual training of the target noise distribution. This flexibility allows us to generate synthetic noisy \& clean image pairs by leveraging diverse guidance noisy images where the denoiser fails to remove noise.

The GuidNoise framework is trained by a novel training method, named as guidance-aware affine feature modification (GAFM), to embed fine-grained noise distribution within the diffusion process.
In addition, we introduce a noise-aware refine loss that improves the diffusion process, enhancing its capability to generate realistic noise distributions while maintaining adaptation capacity from the guidance pair.
Our approach maintains the noisy image quality of previous methods in synthesizing complex noise patterns while overcoming their constraints to enable better generalization across diverse datasets and scenarios. Our contributions are summarized as follows:
\begin{itemize}
    \item We propose GuidNoise, a single-pair guided diffusion model generating realistic noisy images without relying on camera metadata with using a single guidance noise\&clean pair, at an inference phase.
    \item We propose Guidance-aware Affine Feature Modification (GAFM), an approach to capture fine-grained noise within diffusion models using guidance pairs.
    \item We introduce a noise-aware refinement loss that enhances the diffusion process and improves its ability to generate realistic noise distributions.
    \item Our method demonstrates generalized noise synthesis performance across different noisy datasets while maintaining high-quality noise synthesis.
\end{itemize}
\section{Related Works}
\label{sec:related_works}
\paragraph{Noisy Image Synthesis.}
The constraints of real-world noisy datasets have driven efforts towards generative noise modeling, aiming to overcome the dependency on real-world datasets by learning to synthesize realistic noise distributions. 
Generative Adversarial Networks (GAN), Normalizing Flows (NF), and diffusion models have emerged as prominent approaches in this area.
Using synthesized noise from the generative model has shown the potential to improve the denoising networks~\cite{Finetune_w_PNGAN:cai2021learning}.

DANet~\cite{DANet_AKLD_KLD_PGap:yue2020dual} simultaneously generates and removes noise, learning the joint distribution between clean and noisy image pairs from the target scene.
CycleISP~\cite{CycleISP:zamir2020cycleisp} {designs} the {bidirectional} camera imaging pipeline, generating realistic image pairs for RAW and sRGB denoising.
C2N~\cite{C2N:jang2021c2n} employs adversarial loss using unpaired noisy-clean images, which allows for training without relying on paired data. However, its modeling from unpaired data often suffers from limitations in noise quality, such as artifacts and color-shift problems. 
PNGAN~\cite{PNGAN:cai2021learning} tackles the noisy image generation task by breaking it down into image-domain and noise-domain alignment problems using pixel-level modeling. 
Flow-sRGB~\cite{Flow-sRGB:kousha2022modeling} extends the normalizing flow-based NoiseFlow~\cite{NoiseFlow:abdelhamed2019noise}  to the sRGB space, offering improved noise modeling across different image formats. 
NeCA~\cite{NeCA:fu2023srgb} explicitly {consists of} multiple specialized noise networks based on neighbor-noise correlations.
NAFlow~\cite{NAFlow:kim2023srgb} builds upon this by mapping noise from various camera models into a unified latent space.

Diffusion models~\cite{DDPM_Diffusion:ho2020denoising, DDIM_Diffusion:songdenoising, Pred-V_Diffusion:salimansprogressive} have gained significant interest for their ability to {represent} complex distributions without the mode collapse observed in GAN, and without the restrictive invertibility requirements of NF models. Unlike NF models, which require invertible transformations and struggle with high-dimensional complex distributions, diffusion models provide the flexibility to better capture intricate noise patterns. RNSD~\cite{wu2025realistic} is one of the first studies demonstrating the effectiveness of diffusion models in handling the complexities of real-world noise.
Additionally, revisiting the dataset preparation pipeline is recently proposed~\cite{li2025noise}.
{GuidNoise generates high-quality noisy images without requiring camera metadata or a paired noisy image corresponding to a target clean image, thereby enhancing generalizability.}
\begin{figure*}[hbt!]
    \centering
    \begin{subfigure}[b]{0.5\textwidth}
        \centering
        \includegraphics[width=\textwidth]{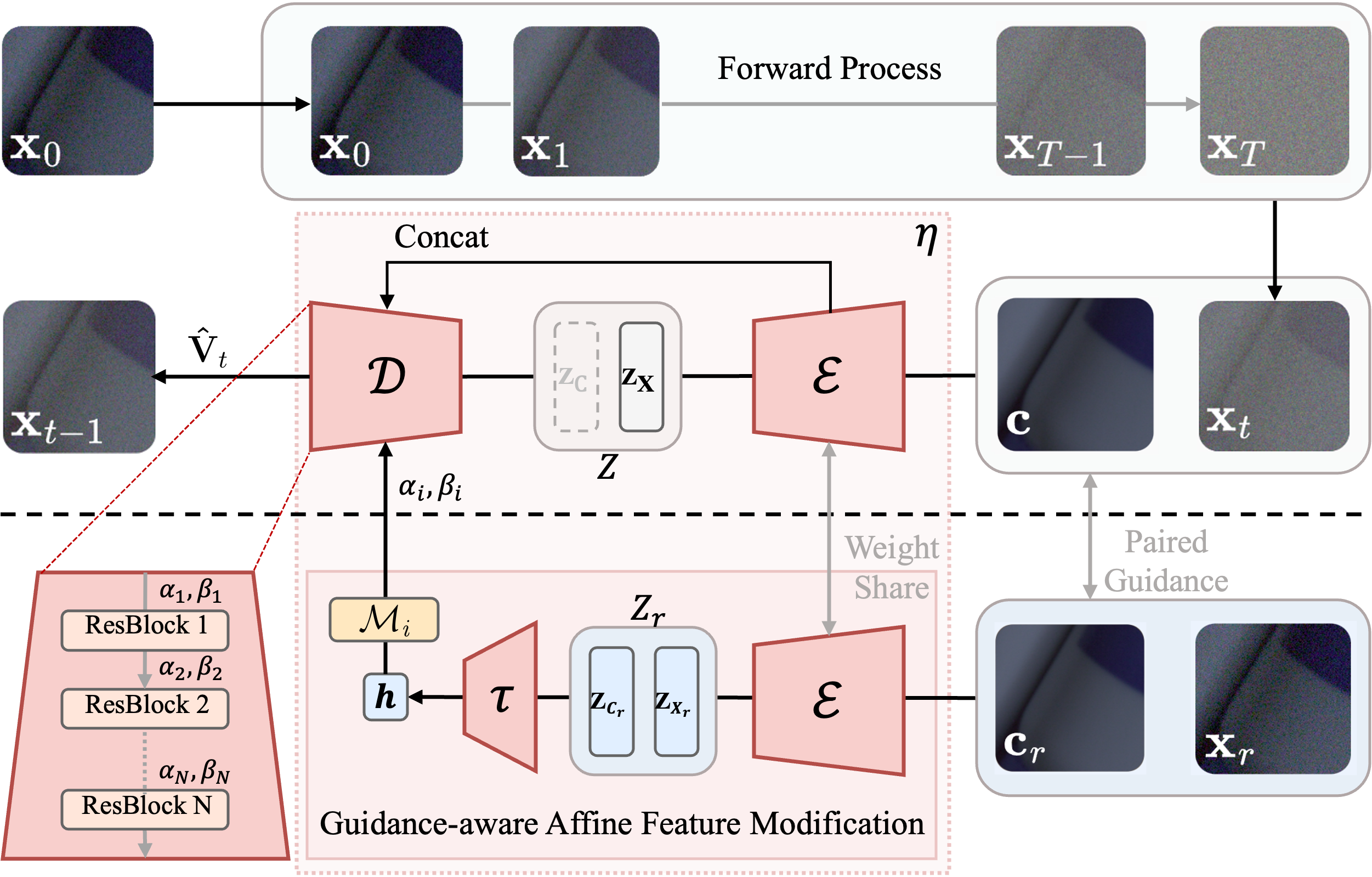}
        \caption{Training}
    \end{subfigure}
    \hspace{1em}
    \begin{subfigure}[b]{0.47\textwidth}
        \centering
        \includegraphics[width=\textwidth]{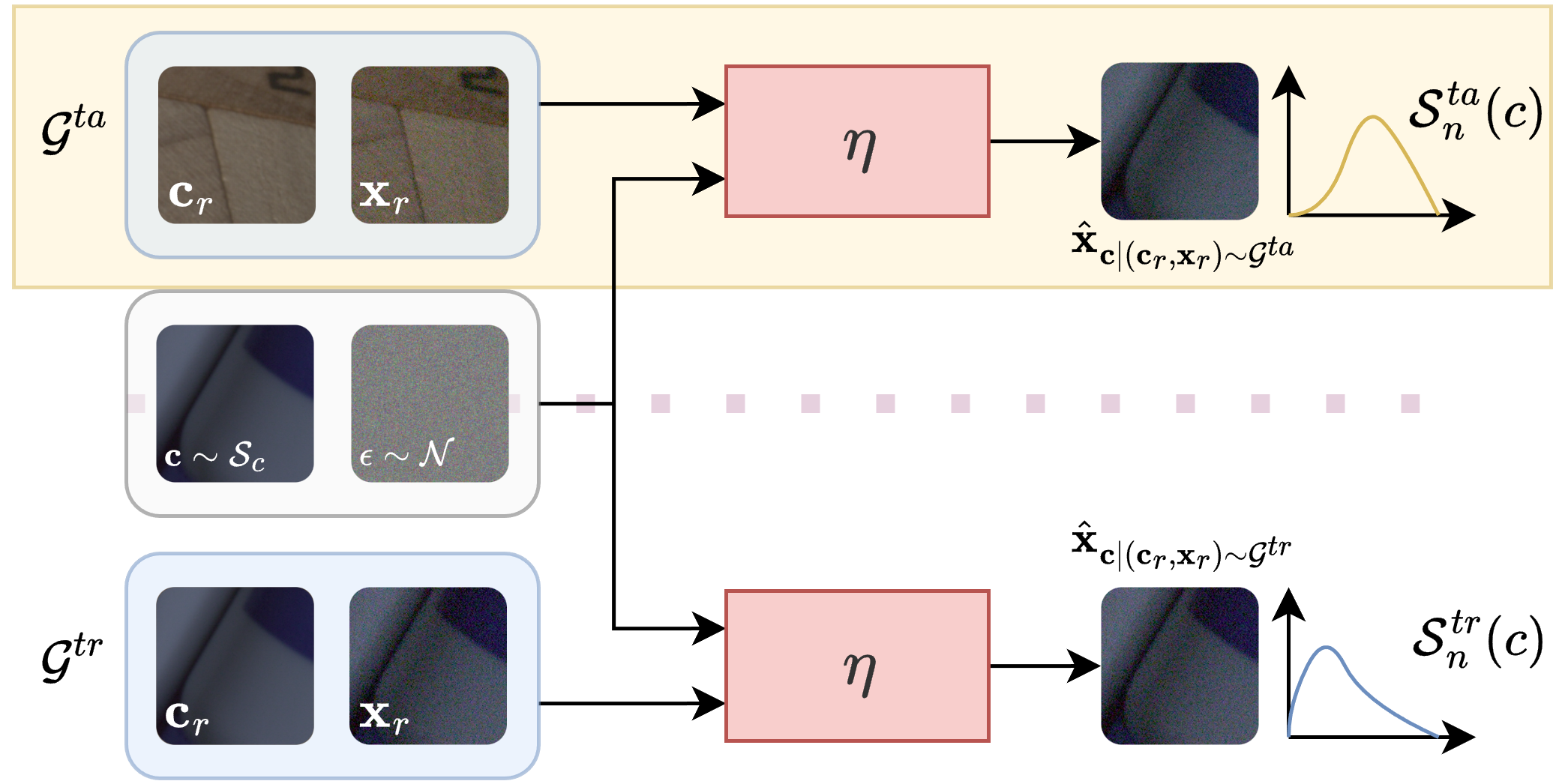}
        \caption{Inference}
    \end{subfigure}
    \caption{\textbf{Overview of the proposed method.} The figure shows the training and inference pipelines of \textit{GuidNoise}. The generation model $\eta(\cdot)$ synthesizes noisy images using decoder $\mathcal{D}$, which takes latent feature $\mathbf{Z_x}$ as input. Each residual block processes a concatenation of the previous decoder output and two encoder features, modulated by affine parameters $(\alpha_i, \beta_i)$.}
    \label{fig:method}
\end{figure*}
\section{Proposed Method}
\subsection{Preliminaries}
\paragraph{Noise Modeling.}
Image noise arises from multiple sources during the imaging process, including the physical limitations of camera sensors and various steps of image processing. Noisy image $\mathbf{x}$ can be defined as a combination of a clean image $ \mathbf{c}$ and noise $\mathbf{n}$ from noise space $\mathcal{S}_n$:
\begin{equation}
    \mathbf{x} = \mathbf{c} + \mathbf{n}, \mathbf{n} \sim\mathcal{S}_n(\mathbf{c}).
\label{eq:noise_modeling}
\end{equation}

\paragraph{Diffusion Model.}
The core idea of diffusion model~\cite{DPM_Diffusion:sohl2015deep} involves gradually adding noise to an image through a forward diffusion process and then learning a reverse process to reconstruct the original image. 
Especially, we leverage Denoising Diffusion Probabilistic Models (DDPM)~\cite{DDPM_Diffusion:ho2020denoising} for diffusion modeling. In the DDPM, the forward process progressively corrupts a target image with Gaussian diffusion noise, as:
\begin{align}
\mathbf{x}_t = & \sqrt{\alpha_t}\mathbf{x_0} + \sqrt{1-\alpha_t} \epsilon, &\text{where} \ \epsilon \sim \mathcal{N}(\mathbf{0}, \mathbf{I}),
\label{eq:diffusion_foward}
\end{align}

To reconstruct the original image from the noisy counterpart, DDPM directly predicts the noise component added during the forward process with two subsequent improvements.
Denoising Diffusion Implicit Models (DDIM)~\cite{DDIM_Diffusion:songdenoising} and v-prediction~\cite{Pred-V_Diffusion:salimansprogressive}. DDIM provides a deterministic sampling,
\begin{align}
\mathbf{x}_{t-1} &= \sqrt{{\alpha}_{t-1}} \left( \frac{\mathbf{x}_t - \sqrt{1 - \alpha_t} \mathbf{\epsilon}_\theta(\mathbf{x}_t, t)}{\sqrt{\alpha}_t} \right) \nonumber \\
&\quad + \sqrt{1 - \alpha_{t-1} - \sigma_t^2} \mathbf{\epsilon}_\theta(\mathbf{x}_t, t) + \sigma_t\epsilon,
\label{eq:ddim_sampling}
\end{align} 
where $ \alpha_t $ and $ \sigma_t $ are time-dependent coefficients. $\alpha_t$ balances the diffusion noise and the target image and $\sigma_t$ balances between deterministic and stochastic sampling.
The v-prediction~\cite{Pred-V_Diffusion:salimansprogressive} estimates the $ \mathbf{v}_t $ instead of predicting the diffusion noise $\epsilon$, which improves stability even with fewer sampling steps~\cite{kingma2023understanding}. This provides a more nuanced combination of the diffusion noise and the target image, allowing better control over the backward process, which defined as:
\begin{equation}
\mathbf{v}_t = \sqrt{\alpha_t} \epsilon - \sqrt{1-\alpha_t} \mathbf{x}_0.
\label{eq:v_prediction}
\end{equation} The neural network $\mathbf{v}_\theta$ is then trained to predict $ \mathbf{v}_t $, following the loss function:
\begin{equation}
    \mathcal{L}_{\text{Diffusion}} = \mathbb{E} \left[ \| \mathbf{v}_\theta(\mathbf{x}_t, t) - \mathbf{v}_t \|^2 \right].
\label{eq:loss_diffusion}
\end{equation}
Then $\mathbf{\epsilon}_\theta(\mathbf{x}_t, t)$ in Eq. (\ref{eq:ddim_sampling}) is reparameterized  by $\mathbf{v}_\theta(\mathbf{x}_t, t)$ in Eq. (\ref{eq:loss_diffusion}).
This enables a diffusion model to estimate the next step more directly with fewer sampling steps.

\subsection{Problem Formulation}
\paragraph{Noise Synthesis given a Single Guidance Pair.} Given clean data space $\mathcal{S}_{c}$ and its noise space $\mathcal{S}_{n}$, our goal is to design a diffusion-based noisy image generator $\eta(\cdot)$ as in,
\begin{equation}
\hat{\mathbf{x}} = \eta(\mathbf{c} | (\mathbf{x}_\text{r}, \mathbf{c}_\text{r})),
\label{eq:definition}
\end{equation} 
where $\mathbf{c} \in \mathcal{S}_c$ is the clean image to which noise is added, and $(\mathbf{x}_\text{r}, \mathbf{c}_\text{r}) \in \mathcal{G}$ is a guidance noisy-clean image pair, with $\mathcal{G} = \{ (\mathbf{x}, \mathbf{c}) \mid \mathbf{x} \sim \mathcal{S}_n(\mathbf{c}),\ \mathbf{c} \in \mathcal{S}_c \}$. Our resultant synthetic noisy image $\hat{\mathbf{x}}$ mimics the real noisy counterpart $\mathbf{x}$ of the input $\mathbf{c}$. Here, we assume that the noise $\mathbf{n}_{x} = \hat{\mathbf{x}} - \mathbf{c} $ and $\mathbf{n}_{r} = \mathbf{x}_\text{r} - \mathbf{c}_\text{r}$ both lie in the same  noise space $\mathcal{S}_{n}$.
In our proposed synthesize phase, \textbf{single-pair
guided noise synthesis diffusion model}, we use the generator $\eta(\cdot)$ trained priorly by the training data set, $\mathcal{S}^{\text{tr}}_{n}$, having different data and noise domains than our target domains $\mathcal{S}^{\text{ta}}_{n}$. 
However, by using the guidance pair $(\mathbf{x}_\text{r}, \mathbf{c}_\text{r})\sim\mathcal{G}^{\text{ta}}$ and given clean image $c\in\mathcal{S}^\text{ta}_{c}$, our noisy image generator $\eta(\cdot)$ generates noise in target space $\mathcal{S}^{\text{ta}}_{n}$.
This demonstrates the strong domain adaptation capability of our noise generator. Notably, our method can generate target-domain noise $\mathcal{S}^{ta}_n$ from a clean image $\mathbf{c}$ from any domain, including $\mathcal{S}^{tr}_c$, using just a single guidance pair, further highlighting its robustness across domains.

\subsection{Single-Pair Guided Noise Synthesis Diffusion}
Our noise generation model $\eta(\cdot)$ is based on a conditional U-Net architecture from DDPM. 
In training phase,  $\eta(\cdot)$ takes $\mathbf{x}_t$ from forward process as an input to backward process, along with a clean image $\mathbf{c}\in\mathcal{S}^{tr}_{c}$ and a guidance pair $(\mathbf{x}_\text{r}$, $\mathbf{c}_\text{r})\in\mathcal{G}^{\text{tr}}$ as additional inputs for guiding the synthetic noisy image generation, as illustrated in Figure~\ref{fig:method}.

\paragraph{Cascade Decoding Architecture.} To implement the architecture, we design a cascade decoding architecture that modifies the original U-Net architecture, forcing each input noise information from the guidance pair to affect different levels of the generation process. 
Therefore, all the input information is embedded by the same encoder $\mathcal{E}$ in parallel. 
The encoder $\mathcal{E}$ consists of multiple encoder blocks $\mathcal{E}_i$, each computing a feature $\mathbf{f}_{\mathbf{x}_t,i}$ for an input image $\mathbf{x}_t$ from the previous feature $\mathbf{f}_{\mathbf{x}_t,i-1}$ and time embedding $\mathbf{t}$ from time step $t$. The entire encoding process can be summarized as
\begin{equation}
\begin{aligned}
\mathbf{f}_{\mathbf{x}_t,i} &= \mathcal{E}_i(\mathbf{f}_{\mathbf{x}_t,i-1}, \mathbf{t}), \ \mathbf{f}_{\mathbf{x}_t,0}= \mathbf{x}_t,\ 1\leq i \leq N, \\
\mathbf{z}_{\mathbf{x}_t} &= \mathbf{f}_{\mathbf{x}_t,N} := \mathcal{E}(\mathbf{x}_t, \mathbf{t}), \ \mathbf{F_{\mathbf{x}_t}} = \{ \mathbf{f}_{\mathbf{x}_t,1}, \mathbf{f}_{\mathbf{x}_t,2}, \dots, \mathbf{f}_{\mathbf{x}_t,N} \}.
\label{eq:encoder}
\end{aligned}
\end{equation}
Here, $\mathbf{z_{x_t}}$ is the final embedding and $\mathbf{F_{x_t}}$ is the set of intermediate features. 

\paragraph{Guidance-aware Affine Feature Modification.}
Inspired by the conditional feature modification methods~\cite{perez2018film,dhariwal2021diffusion,wu2025realistic}, applying the affine transform of the features given the conditional information, we propose a \textbf{guidance-aware affine feature modification} method to embed the domain-specific noise information from the guidance pair ($\mathbf{x}_\text{r}$, $\mathbf{c}_\text{r}$).
Here, the embedding $\mathbf{z_r}$ from guidance image pair ($\mathbf{x}_\text{r}$, $\mathbf{c}_\text{r}$) is an input into the Noise-aware guidance module $\tau$ that processes $\mathbf{z_r}$ to form the guidance embedding $\mathbf{h}$,
\begin{equation}
     \mathbf{h} = \tau(\mathbf{z}_\text{r},\mathbf{t}), \ \mathbf{z}_\text{r} = \text{concat}(\mathcal{E}(\mathbf{x}_\text{r}, t),\mathcal{E}(\mathbf{c}_\text{r}, t)).
     \label{eq:guidance_module}
\end{equation}

The concatenated guidance embedding, combined with the time embedding $\mathbf{t}$, serves as guidance input to the decoder $\mathcal{D}$. 
Note that the decoder $\mathcal{D}$ differs from the encoder $\mathcal{E}$, where the encoder does not use any condition, such as the guidance embedding.
By conveying the guidance through $\tau(\cdot)$, we prevent the direct use of raw guidance features in the decoding process. Instead, $\tau(\cdot)$ focuses on capturing the essential characteristics of the noise distribution, which enhances generalization to diverse domains. 
Specifically, letting $\mathcal{D}_{i,j}$, be the $j$-th layer in the $i$-th decoder block and $\mathcal{M}_{i}(\mathbf{h}, \mathbf{t})$ be a multilayer perceptron (MLP) to convey guidance in $t$ to the $i$-th decoder block $\mathcal{D}_{i}$ in the form of an affine coefficient $\alpha$ and $\beta$, the entire decoding process is given as:
\begin{align}
    &\alpha_i, \ \beta_i = \mathcal{M}_i(\mathbf{h}, \mathbf{t}), \nonumber \\
    &\mathbf{g}_{i,j} =(1+\alpha_i) \mathcal{D}_{i, j}(\text{concat}(\mathbf{g}_{i,j-1}, \mathbf{f}_{\mathbf{x}_t, i}, \mathbf{f}_{\mathbf{c}, i})) + \beta_i, \nonumber \\ & \ \ \ \ \ 1 \leq j \leq L-1, 1 \leq i \leq N, \label{eq:decoder} \\
    &\mathbf{g}_{i, L} =\mathcal{D}_{i,L}(\mathbf{g}_{i-1, L}, \mathbf{f}_{\mathbf{x}_t, i-1}, \mathbf{f}_{\mathbf{c}, i-1}, \mathbf{h}, \mathbf{t}), \nonumber 
\end{align}
where 
$\mathbf{g}_{1,0}=\mathbf{z}_{\mathbf{x}_t}$  and  
$\mathbf{g}_{i \ge 2, 0}=\mathbf{g}_{i-1,L}$. Then the final decoder output of the backward process becomes $\mathbf{g}_{N, L}$. Letting the backward process be $\eta(\cdot)$ as in Eq. (\ref{eq:loss_diffusion}), $\mathbf{g}_{N, L}$ for $\mathbf{x}_t$ becomes 
${\hat{\mathbf{x}}}_{t}$ that can be expressed by
\begin{equation}
\begin{aligned}
    {\hat{\mathbf{x}}}_{t} &= \eta(\mathbf{x}_t, \mathbf{c}, \mathbf{x}_\text{r}, \mathbf{c}_\text{r},  t) \label{eq:prediction1} \\
    &= \mathcal{D}(\mathbf{z_{x_t}}, \mathbf{F_{x_t}}, \mathbf{F_{c}}, \mathbf{h}, \mathbf{t}).
\end{aligned}
\end{equation} 
Finally ${\mathbf{x}}_{t-1}$ is obtained from Eq. (\ref{eq:ddim_sampling}) by reparameterizing  $\mathbf{\epsilon}_\theta$ to $\eta$ in Eq. (\ref{eq:prediction1}). The model is primarily optimized by the diffusion loss in Eq. (\ref{eq:loss_diffusion}). Note that $\mathbf{v}_\theta(\cdot)$ in Eq. (\ref{eq:loss_diffusion}) is replaced by $\eta(\cdot)$, i.e., $\mathbf{h}, \mathbf{x}_\text{r}, \mathbf{c}_\text{r}$ are newly explored in our model. 

\paragraph{Refine Loss for Preserving Noise Components.}
As previously mentioned, the standard diffusion loss in Eq.~(\ref{eq:loss_diffusion}) relies on $L_2$ normalization, which theoretically suppresses high-frequency components—where most noise information resides. To address this issue, we introduce a refinement loss into the backward diffusion process, ensuring distribution alignment between the final synthesized noisy image and the ground truth noisy image.

Specifically, during the sampling process from T to 0, we track the gradients for the last $T_{\text{split}}$ steps, where most of the fine-grained noise details are synthesized. The refine loss aims to improve the similarity between the histogram of the synthesized  noisy image and the ground truth noisy image. 
To this end, we use Kullback-Leibler Divergence (KLD) $D_{\text{KL}}$ and differentiable histogram~\cite{HistLoss:UstinovaNIPS16} $\bar{H}$.
The refine loss is given by 
\begin{equation}
    \mathcal{L}_{\text{refine}} = D_{\text{KL}}(\bar{H}(\hat{\mathbf{x}}) \parallel \bar{H}(\mathbf{x})) + \gamma \mathbb{E} \left[ \| \mathbf{\hat{\mathbf{x}}} - \mathbf{x} \|^2\right],
    \label{eq:refine_loss}
\end{equation}
where $\bar{H}(\hat{\mathbf{x}})$ and $\bar{H}(\mathbf{x})$ represent the differentiable histograms of the synthesized noisy image and the ground truth noisy image, respectively. $\gamma$ is weight parameter of the regularization term. The KL divergence ensures that the distribution of the synthesized noisy image closely matches that of the ground truth by tuning backward process.

By combining the diffusion loss $\mathcal{L}_{\text{Diffusion}}$ in Eq. (\ref{eq:loss_diffusion}) and the refine loss 
$\mathcal{L}_{\text{refine}}$ in Eq. (\ref{eq:refine_loss}), the overall loss is given by 
\begin{equation}
    \mathcal{L} = \mathcal{L}_{\text{Diffusion}} + \lambda \mathcal{L}_{\text{Refine}},
    \label{eq:refine_and_diffusion}
\end{equation}
where $\lambda$ is a weighting factor that controls the contribution of the refine loss.

\paragraph{Inference Phase.}
During inference, the noisy image generator $\eta(\cdot)$ synthesizes a noisy image $\hat{x}$ for a clean image $\mathbf{c} \in \mathcal{S}_c^{ta}$, guided by a single noisy-clean reference pair $(\mathbf{x}_\text{r}, \mathbf{c}_\text{r}) \sim \mathcal{G}^{ta}$. Unlike the training phase, we no longer require access to any ground-truth noisy image corresponding to $\mathbf{c}$. Instead, the model leverages the noise characteristics extracted from $(\mathbf{x}_\text{r}, \mathbf{c}_\text{r})$ and transfers them to the input image $\mathbf{c}$. By leveraging the guidance-aware affine modulation and refine loss, the model adapts effectively to unseen target domains, generating noise patterns that closely resemble the target domain distribution.
\begin{figure*}[!ht]
    \newcommand{\imgwidth}{\textwidth}
    \newcommand{\subfigwidth}{0.22\textwidth}
    \newcommand{\labelfigwidth}{0.05\textwidth}
    \newcommand{\hspacing}{0.1mm}
    \newcommand{\minipagewidth}{0.499\textwidth}
    
    \centering
    \begin{minipage}{\minipagewidth}
        \centering
        \begin{subfigure}[b]{\labelfigwidth}
            \centering
            \caption*{}
        \end{subfigure}%
        \hspace{\hspacing} 
        \begin{subfigure}[b]{\subfigwidth}
            \centering
            \caption*{\scriptsize NeCA-W}
        \end{subfigure}%
        \hspace{\hspacing}
        \begin{subfigure}[b]{\subfigwidth}
            \centering
            \caption*{\scriptsize NAFlow}
        \end{subfigure}%
        \hspace{\hspacing}
        \begin{subfigure}[b]{\subfigwidth}
            \centering
            \caption*{\scriptsize {Ours}}
        \end{subfigure}%
        \hspace{\hspacing}
        \begin{subfigure}[b]{\subfigwidth}
            \centering
            \caption*{\scriptsize Real}
        \end{subfigure}
        \\
        
        \begin{subfigure}[b]{\labelfigwidth}
            \centering
            \caption*{\raisebox{0.6\height}{\rotatebox{90}{\scriptsize SIDD}}}
        \end{subfigure}%
        \hspace{\hspacing}
        \begin{subfigure}[b]{\subfigwidth}
            \centering
            \includegraphics[width=\imgwidth]{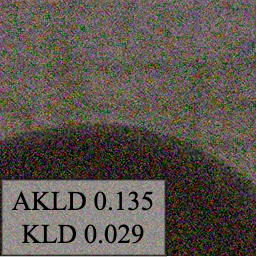}
        \end{subfigure}%
        \hspace{\hspacing}
        \begin{subfigure}[b]{\subfigwidth}
            \centering
            \includegraphics[width=\imgwidth]{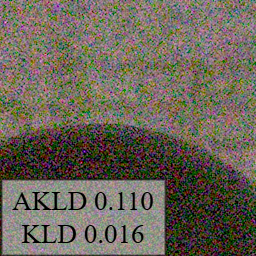}
        \end{subfigure}%
        \hspace{\hspacing}
        \begin{subfigure}[b]{\subfigwidth}
            \centering
            \includegraphics[width=\imgwidth]{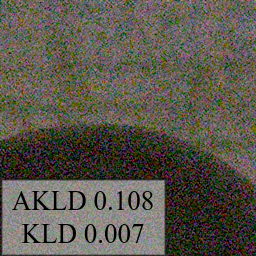}
        \end{subfigure}%
        \hspace{\hspacing}
        \begin{subfigure}[b]{\subfigwidth}
            \centering
            \includegraphics[width=\imgwidth]{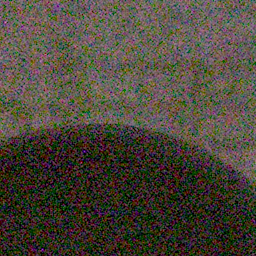}
        \end{subfigure}
        \\
        
        \begin{subfigure}[b]{\labelfigwidth}
            \centering
            \caption*{\raisebox{0.6\height}{\rotatebox{90}{\scriptsize PolyU}}}
        \end{subfigure}%
        \hspace{\hspacing}
        \begin{subfigure}[b]{\subfigwidth}
            \centering
            \includegraphics[width=\imgwidth]{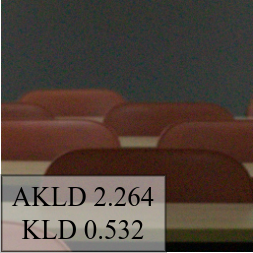}
        \end{subfigure}%
        \hspace{\hspacing}
        \begin{subfigure}[b]{\subfigwidth}
            \centering
            \includegraphics[width=\imgwidth]{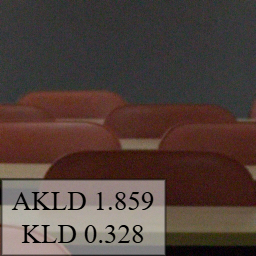}
        \end{subfigure}%
        \hspace{\hspacing}
        \begin{subfigure}[b]{\subfigwidth}
            \centering
            \includegraphics[width=\imgwidth]{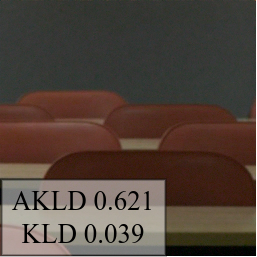}
        \end{subfigure}%
        \hspace{\hspacing}
        \begin{subfigure}[b]{\subfigwidth}
            \centering
            \includegraphics[width=\imgwidth]{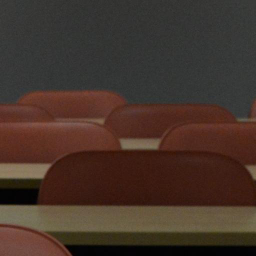}
        \end{subfigure}
    \end{minipage}%
    \begin{minipage}{\minipagewidth}
        \centering
        \begin{subfigure}[b]{\labelfigwidth}
            \centering
            \caption*{}
        \end{subfigure}%
        \hspace{\hspacing}
        \begin{subfigure}[b]{\subfigwidth}
            \centering
            \caption*{\scriptsize NeCA-W}
        \end{subfigure}%
        \hspace{\hspacing}
        \begin{subfigure}[b]{\subfigwidth}
            \centering
            \caption*{\scriptsize NAFlow}
        \end{subfigure}%
        \hspace{\hspacing}
        \begin{subfigure}[b]{\subfigwidth}
            \centering
            \caption*{\scriptsize {Ours}}
        \end{subfigure}%
        \hspace{\hspacing}
        \begin{subfigure}[b]{\subfigwidth}
            \centering
            \caption*{\scriptsize Real}
        \end{subfigure}
        \\
        
        \begin{subfigure}[b]{\labelfigwidth}
            \centering
            \caption*{\raisebox{0.8\height}{\rotatebox{90}{\scriptsize Nam}}}
        \end{subfigure}%
        \hspace{\hspacing}
        \begin{subfigure}[b]{\subfigwidth}
            \centering
            \includegraphics[width=\imgwidth]{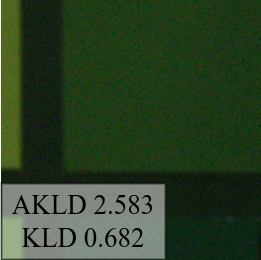}
        \end{subfigure}%
        \hspace{\hspacing}
        \begin{subfigure}[b]{\subfigwidth}
            \centering
            \includegraphics[width=\imgwidth]{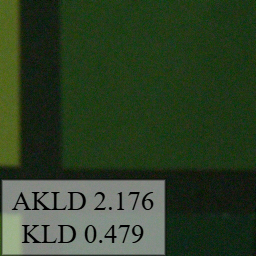}
        \end{subfigure}%
        \hspace{\hspacing}
        \begin{subfigure}[b]{\subfigwidth}
            \centering
            \includegraphics[width=\imgwidth]{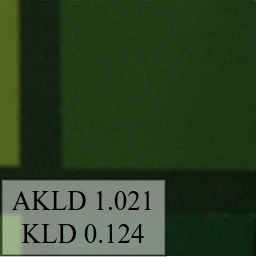}
        \end{subfigure}%
        \hspace{\hspacing}
        \begin{subfigure}[b]{\subfigwidth}
            \centering
            \includegraphics[width=\imgwidth]{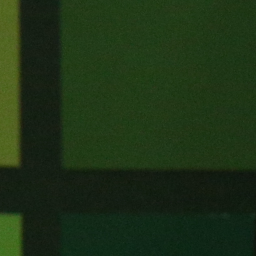}
        \end{subfigure}
        \\
        
        \begin{subfigure}[b]{\labelfigwidth}
            \centering
            \caption*{\raisebox{0.5\height}{\rotatebox{90}{\scriptsize SIDD+}}}
        \end{subfigure}%
        \hspace{\hspacing}
        \begin{subfigure}[b]{\subfigwidth}
            \centering
            \includegraphics[width=\imgwidth]{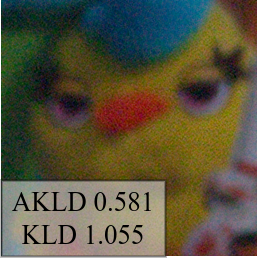}
        \end{subfigure}%
        \hspace{\hspacing}
        \begin{subfigure}[b]{\subfigwidth}
            \centering
            \includegraphics[width=\imgwidth]{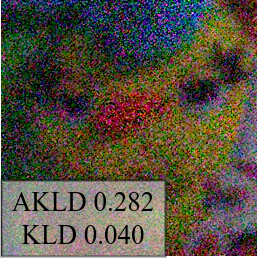}
        \end{subfigure}%
        \hspace{\hspacing}
        \begin{subfigure}[b]{\subfigwidth}
            \centering
            \includegraphics[width=\imgwidth]{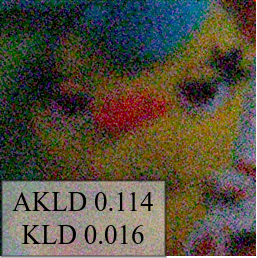}
        \end{subfigure}%
        \hspace{\hspacing}
        \begin{subfigure}[b]{\subfigwidth}
            \centering
            \includegraphics[width=\imgwidth]{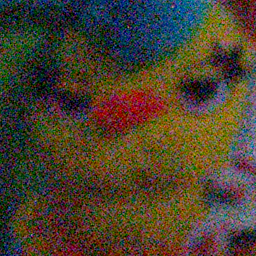}
        \end{subfigure}
    \end{minipage}
    \caption{\textbf{Qualitative comparison of synthetic noisy imagess} on SIDD-Validation (SIDD), SIDD+, PolyU and Nam dataset.}
    \label{fig:noise_quality}
\end{figure*}
\begin{table*}
  \centering
    \resizebox{0.95\textwidth}{!}{
      \begin{tabular}{c r | cc cc cc c }
        \toprule
        \multirow{1}{*}{Camera} & 
        \multirow{1}{*}{Metrics} & 
        \multirow{1}{*}{C2N} &
        \multirow{1}{*}{Flow-sRGB} &
        \multirow{1}{*}{NeCA-S} &
        \multirow{1}{*}{NeCA-W} &
        \multirow{1}{*}{NAFlow} &
        \textbf{\multirow{1}{*}{Ours}}\\
        \midrule
    
        \multirow{1}{*}{G4} &
        KLD / AKLD &
        0.098 / 0.194 & 
        0.123 / 0.229& 
        0.403 / 1.680& 
        0.043 / 0.153& 
        0.025 / 0.137& 
        \textbf{0.018} / \textbf{0.120}\\ 

        \multirow{1}{*}{GP} &
        KLD / AKLD &
        0.693 / 0.428& 
        0.076 / 0.140& 
        0.125 / 0.362& 
        0.043 / 0.126& 
        0.035 / 0.118& 
        \textbf{0.010} / \textbf{0.111}\\ 
    
        \multirow{1}{*}{IP} &
        KLD / AKLD&
        0.078 / 0.230& 
        0.077 / 0.216& 
        0.311 / 1.205& 
        0.048 / 0.116& 
        0.034 / 0.152& 
        \textbf{0.017} / \textbf{0.110}\\
    
        \multirow{1}{*}{N6} &
        KLD / AKLD&
        0.416 / 0.293 & 
        0.148 / 0.179& 
        0.282 / 1.058& 
        0.055 / 0.130& 
        0.031 / 0.111& 
        \textbf{0.013} / \textbf{0.108}\\ 

        \multirow{1}{*}{S6} &
        KLD / AKLD&
        0.683 / 0.418& 
        0.109 / 0.184& 
        0.171 / 0.462& 
        0.052 / 0.194& 
        0.027 / 0.136& 
        \textbf{0.013} / \textbf{0.117}\\
    
        \midrule
        \multirow{1}{*}{Avg} &
        KLD\text{$\downarrow$} / AKLD\text{$\downarrow$}&
        0.394 / 0.313& 
        0.112 / 0.193& 
        0.259 / 0.953& 
        0.048 / 0.144& 
        0.031 / 0.131& 
        \textbf{0.014} / \textbf{0.113}\\
        \bottomrule
      \end{tabular}
    }
  \caption{
  \textbf{Quantitative comparison of synthetic noise} on SIDD-Validation dataset. }
  \label{tab:noise_quality_comparison_main}
\end{table*}

\section{Experiments}
\subsection{Experimental Setup}
\label{sec:exp_setup}
\paragraph{Evaluation.}
We compare the similarity of noisy images with several noise synthesis methods: C2N~\cite{C2N:jang2021c2n}, Flow-sRGB~\cite{Flow-sRGB:kousha2022modeling}, NeCA-S/NeCA-W~\cite{NeCA:fu2023srgb}, and NAFlow~\cite{NAFlow:kim2023srgb}.
{We note that C2N, Flow-sRGB, and NAFlow, as lightweight models, exhibit performance limitations, whereas NeCA includes models with varying parameter sizes, as detailed in the Supplementary Material. The parameter size of the noise synthesis module in our method is of the same order as NeCA. Also, while other synthesis methods have fewer parameters, they demonstrate inferior synthesis quality.}
To evaluate the quality of synthetic noisy images, we use KLD and AKLD~\cite{DANet_AKLD_KLD_PGap:yue2020dual} as metrics. Furthermore, we train DnCNN~\cite{DnCNN:zhang2017beyond} from scratch using synthetic datasets to indirectly evaluate the synthetic noisy images. We then train NAFNet~\cite{NAFNet:chen2022simple} via a self-augmentation to evaluate enhanced denoising performance. For denoising evaluation, we use PSNR and SSIM metrics.

\paragraph{Dataset.}
We primarily use the real-world noise dataset SIDD~\cite{SIDD:SIDD_2018_CVPR}. For {GuidNoise}, we use the SIDD-Medium dataset split into 256 × 256 patches. To train DnCNN from scratch, we use subset of SIDD-Medium dataset split into 512 × 512 patches, following the experimental settings in ~\cite{NeCA:fu2023srgb} and ~\cite{NAFlow:kim2023srgb}.
We evaluate synthesized noisy images using SIDD-Validation~\cite{SIDD:SIDD_2018_CVPR}. To show our model's generalized performance, we use SIDD+~\cite{SIDD+:abdelhamed2020ntire}, PolyU~\cite{PolyU:xu2018real}, and Nam~\cite{NAM:nam2016holistic} for image synthesis, and SIDD-Benchmark\footnote{Our results may slightly differ from prior SIDD benchmarks due to its recent migration to a Kaggle competition.}~\cite{SIDD:SIDD_2018_CVPR} for image denoising. We crop PolyU and Nam datasets to 512 × 512 pixels.
For self-augmentation, we sample the training set from the SIDD-Validation using proportions of 1/16, 1/8, 1/4, and 1/2. with the remaining one as test set.
\begin{table}[!t]
    \centering
    \resizebox{0.475\textwidth}{!}{
    \begin{tabular}{l | cc cc  cc }
    \toprule 
    \multirow{2}{*}{Method} &  SIDD+ & PolyU & Nam \\
    \cmidrule{2-4}
    & KLD\text{$\downarrow$} / AKLD\text{$\downarrow$} & KLD\text{$\downarrow$} / AKLD\text{$\downarrow$}  & KLD\text{$\downarrow$} / AKLD\text{$\downarrow$} \\ \midrule
    C2N & 0.192 / 0.302 & 0.627 / 2.399 & 0.484 / 2.057 \\
    NeCA-S & 0.298 / 1.301 & 1.309 / 3.239  & 1.161 / 3.144 \\
    NeCA-W & 0.174 / 0.207 & 0.139 / 0.795 & \textbf{0.141} / 0.542 \\
    NAFlow   & \textbf{0.049} / 0.291 & 0.348 / 1.845  & 0.456 / 2.040 \\
    \midrule
    Ours    & 0.050 / \textbf{0.176} & \textbf{0.115} / \textbf{0.587}  & 0.153 / \textbf{0.414} \\
    \bottomrule
    \end{tabular}
    }
    \caption{\textbf{Quantitative comparison of noise similarity} across diverse real-world noisy datasets.}
\label{tab:noise_quality_comparision_datasets}
\end{table}

\paragraph{Training.}
{GuidNoise} is trained for 300K iterations using the diffusion loss in Eq.~(\ref{eq:loss_diffusion}), with a learning rate of $1\mathrm{e}{-4}$ and a batch size of 4.
It is then refined for 50K iterations with a learning rate of $1\mathrm{e}{-5}$ and a batch size of 1, using both the diffusion loss and the refining loss as defined in Eq.~(\ref{eq:refine_and_diffusion}). All optimization in {GuidNoise} is performed using AdamW~\cite{AdamW:loshchilov2018decoupled}. We set $\lambda{=}1$, $\gamma{=}0.1$, $T_{\text{split}}{=}2$, and $T{=}50$. To train DnCNN~\cite{DnCNN:zhang2017beyond} from scratch, we base our approach on the experimental settings in ~\cite{NeCA:fu2023srgb} and ~\cite{NAFlow:kim2023srgb}. We use a learning rate of $1\mathrm{e}{-3}$ and run the training for 300 epochs, with a batch size of 32 and random cropping. For optimization, we employ the Adam optimizer~\cite{Adam:kingma2014adam}.
To train NAFNet from scratch via self-augmentation, we use a learning rate $\eta_0$ of $1\mathrm{e}{-3}$  with an exponential scheduler $(\eta_x = \max(0.95^x \cdot \eta_0, 1 \times 10^{-6}))$, which decays the learning rate  by 0.95 every 100 iterations initially, then every 1K iterations. We use a batch size of 16 and the PSNR Loss from NAFNet experiments~\cite{NAFNet:chen2022simple} for 10K iterations.

\begin{table*}[!t]
  \centering
  \aboverulesep=0ex
    \belowrulesep=0ex
    \resizebox{0.8\textwidth}{!}{
      \begin{tabular}{c | c | c c c c c c | c}
        \toprule
        \multirow{1}{*}{Dataset} & 
        \multirow{1}{*}{Metrics} & 
        \multirow{1}{*}{C2N} &
        \multirow{1}{*}{Flow-sRGB} &
        \multirow{1}{*}{NeCA-S} &
        \multirow{1}{*}{NeCA-W} &
        \multirow{1}{*}{NAFlow} &
        \textbf{\multirow{1}{*}{Ours}} &
        \multirow{1}{*}{Real} \\
        \midrule
    
        SIDD- & 
        PSNR\text{$\uparrow$} &
        33.72 & 
        32.83 & 
        35.02 & 
        36.20 & 
        37.00 & 
        \textbf{37.07} & 
        37.16 \\ 
        
        Validation & 
        SSIM\text{$\uparrow$} &
        0.815 & 
        0.861 & 
        0.880 & 
        0.897 & 
        0.895 & 
        \textbf{0.901} & 
        0.899 \\ 
        \midrule
    
        SIDD- & 
        PSNR\text{$\uparrow$} &
        35.05 & 
        33.88 & 
        35.46 & 
        36.61 & 
        37.39 & 
        \textbf{37.48} & 
        37.60 \\ 
        
        Benchmark & 
        SSIM\text{$\uparrow$} &
        0.802 & 
        0.849 & 
        0.872 & 
        0.888 & 
        0.889 & 
        \textbf{0.895} & 
        0.890 \\ 
        \bottomrule
      \end{tabular}
    }
  \caption{
  \textbf{Quantitative comparison of denoising performance} on SIDD-Validation and SIDD-Benchmark dataset.}
  \label{tab:denoiser_performance_comparision}
\end{table*}
\subsection{Comparison Results}
\label{sec:comparison}
 We compare the camera-wise similarity of the synthetic SIDD-Validation datasets with several noise synthesis methods in Table~\ref{tab:noise_quality_comparison_main}. 
{GuidNoise} exhibits remarkable AKLD (0.113) score, reduced from NAFlow (0.131) and NeCA-W (0.144). {GuidNoise} also shows KLD (0.014) score, nearly halving score of NAFlow (0.031) and NeCA-W (0.048).
We highlight that C2N is trained on an unpaired dataset. Furthermore, Flow-sRGB results do not include ISO 6400, as the setting is not supported in Flow-sRGB. Therefore, the direct comparisons maybe less suitable.

\begin{figure*}[t]
    \centering
    \begin{subfigure}[b]{0.38\textwidth}
        \centering
        \includegraphics[width=0.99\textwidth]{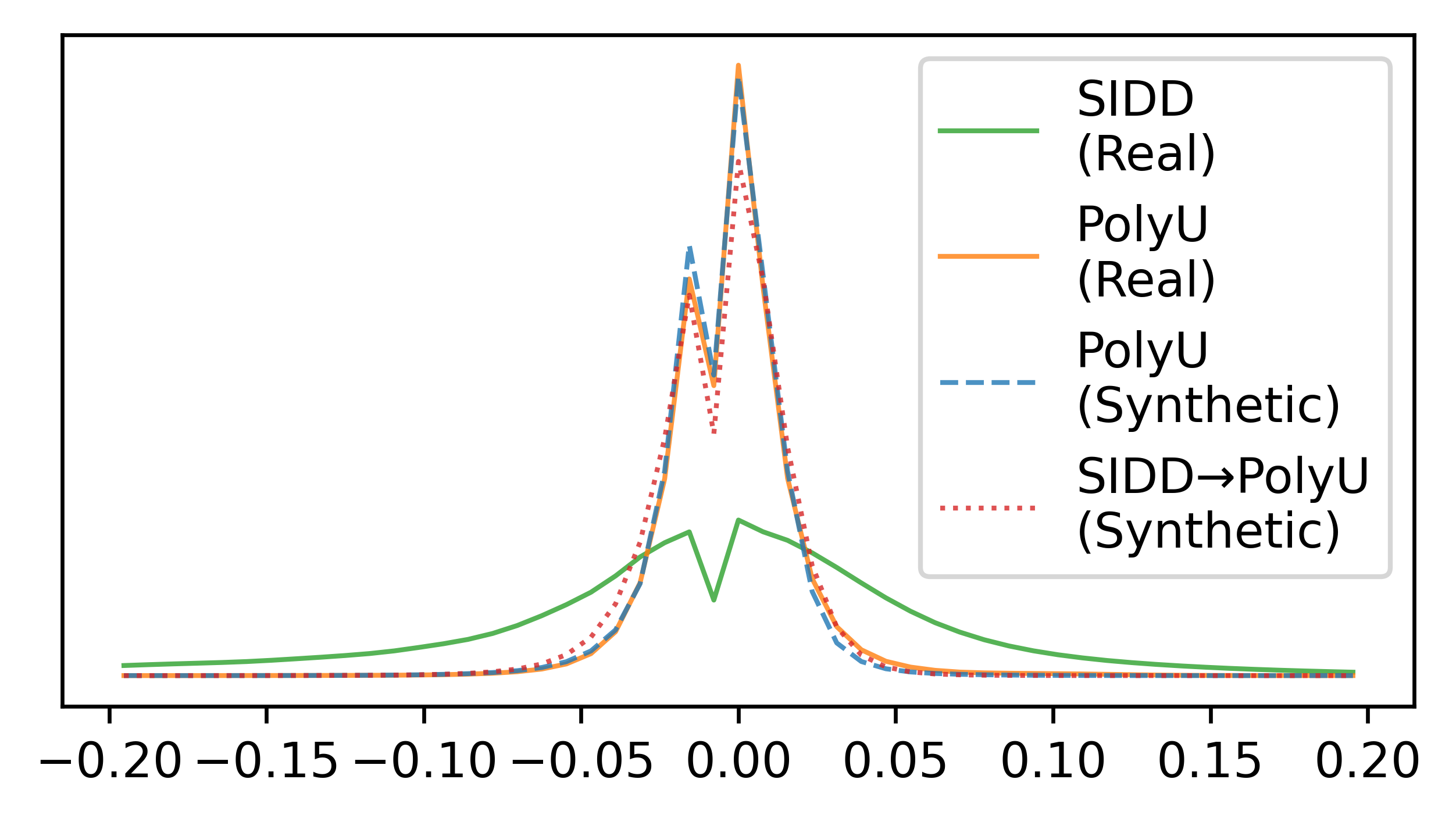} 
        \caption{ Noise distribution from {diverse synthesize cases.}}
        \label{reb:fig:sidd_polyu}
    \end{subfigure}
    \begin{subfigure}[b]{0.3\textwidth}
        \centering
        \includegraphics[width=\textwidth]{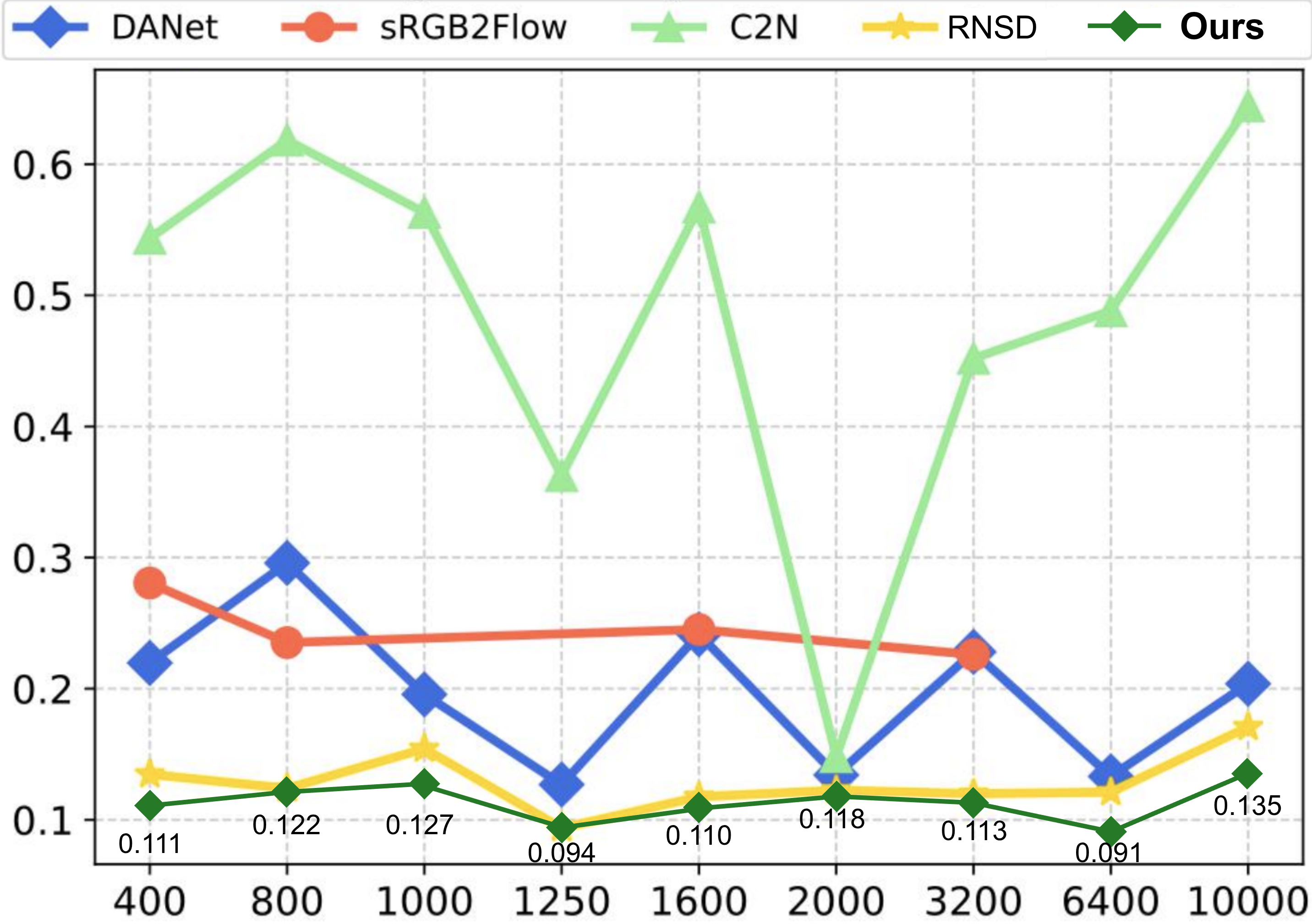}
        \caption{Average AKLD across ISO.}
        \label{fig:compare_iso}
    \end{subfigure}
    \centering
    \begin{subfigure}[b]{0.3\textwidth}
        \centering
        \includegraphics[width=\textwidth]{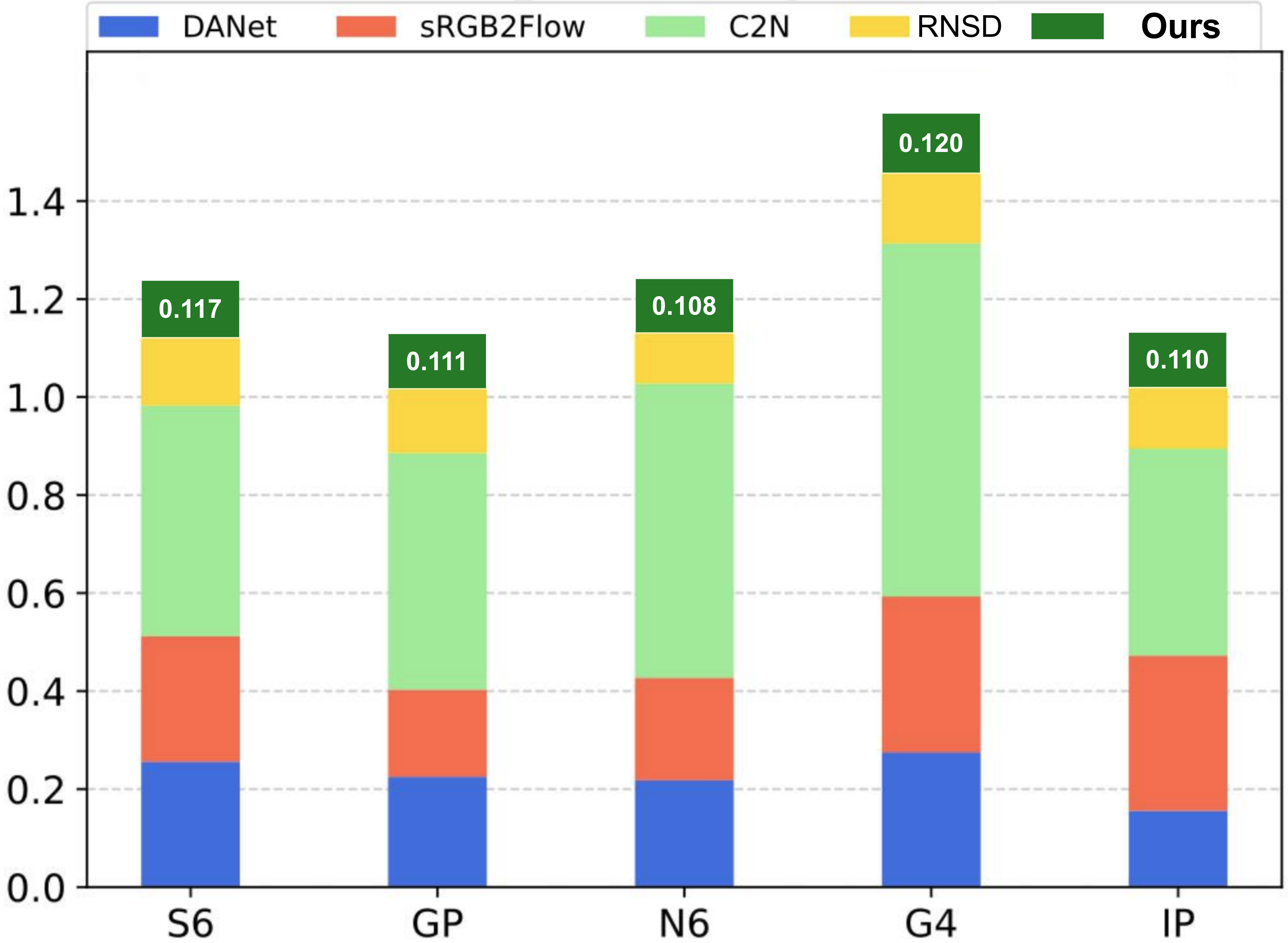}
        \caption{Average AKLD across device.}
        \label{fig:compare_cam}
    \end{subfigure}
    \centering
    \caption{
    \textbf{Comparison with reported performance results} from RNSD across various camera settings.
    }    
    \label{reb:fig:compare_cam_iso}
\end{figure*}

\paragraph{Generalizable Noise synthesis.}
To demonstrate the generalization performance of {GuidNoise}, we synthesize noisy images using SIDD+, PolyU, and Nam datasets, as shown in Table~\ref{tab:noise_quality_comparision_datasets}. Note that NeCA-S/W are evaluated using the average score of each device-specific model. Remarkably, despite being trained solely on smartphone-based SIDD dataset, {GuidNoise} shows significantly improved similarity even on real-world noise datasets captured with DSLR cameras, such as PolyU and Nam. This highlights {GuidNoise}'s generalized performance to capture noise distributions across different camera types without specific training. Specifically, the AKLD reduces from 0.207 (NeCA-W) and 0.291 (NAFlow) to 0.176 on SIDD+, from 0.795 (NeCA-W) to 0.587 on PolyU, and from 0.542 (NeCA-W) to 0.414 on Nam. We present the visualization in Figure~\ref{fig:noise_quality}. 

To further investigate the case with a large gap between two noise maps from different settings in Sensor, ISO, scene, etc., {we additionally} conducted an experimental analysis on \textbf{unpaired synthesis scenarios} (i.e., synthetic noise transfer from guide noisy images in the unseen set (PolyU/Nam) to target noisy images in the trained set (SIDD). 
We measured the noise distributions of the real noises from SIDD training data and synthesized the noises from the PolyU test data in Figure~\ref{reb:fig:sidd_polyu}.
The results show a large gap between SIDD and real PolyU noises, which arises from a setting gap. To reduce the influence of the setting gap during synthesis, we design the feature-level affine transforms in Eq. (8) (i.e., $(1+\alpha) \mathcal{D}+\beta$) and Figure~\ref{fig:method}. Consequently, even in the case of a large noise gap in an unpaired synthesis scenario of clean SIDD input and PolyU guide (SIDD$\rightarrow$PolyU), we achieve close noise distribution to real PolyU. 
While our approach uses only paired noisy-clean images (SIDD) for training, our synthesized noisy images in any setting are valid without any assumption on the correlation between input and guide. 

{In terms of generalizability, RNSD and the proposed {GuidNoise} exhibit distinct characteristics. While RNSD utilizes clean input along with camera conditions, {GuidNoise} instead leverages guide image pairs, eliminating the need for explicit camera condition metadata.}
We added the AKLD results of RNSD to compare our model across different ISO and sensor settings in Figure~\ref{fig:compare_iso} and ~\ref{fig:compare_cam}, where ours shows consistent performance across ISO levels and sensor types, with better similarity in high ISO cases.
Furthermore, our approach shows better performance in PSNR Gap (0.12) than  RNSD (0.54), without using camera metadata.

\paragraph{Comparison in Image denoising on synthetic dataset.} Furthermore, to indirectly compare the similarity of synthesized noisy images to real noisy images across various noise synthesis methods, we train DnCNN using only the synthetic dataset. The results are shown in Table~\ref{tab:denoiser_performance_comparision}. DnCNN trained with {GuidNoise} demonstrates remarkable denoising performance on both the SIDD-Validation and SIDD-Benchmark datasets, achieving the highest PSNR and SSIM values among synthetic methods. 
Specifically, {GuidNoise} achieves a PSNR of 37.07 and SSIM of 0.901 on SIDD-Validation, as well as a PSNR of 37.48 and SSIM of 0.895 on SIDD-Benchmark. This result closely approximates the performance of real noisy images (Real), where PSNR and SSIM reach 37.16 and 0.899 on SIDD-Validation and 37.60 and 0.890 on SIDD-Benchmark, respectively. 
We present the visualization of denoised noisy images in the supplementary material.
Notably, the denoised images from the denoising model trained with synthesized images from {GuidNoise} are visually closest to the clean image. This indirectly supports the realistic synthesis performance of {GuidNoise}.

\subsection{Discussion}
\begin{table}[t]
    \centering
    \aboverulesep=0ex
    \belowrulesep=0ex
    \resizebox{0.45\textwidth}{!}{
    \begin{tabular}{c|c|c|c}
        \toprule
         Synth. PolyU & Real PolyU & Synth. Nam & Real Nam\\
        \midrule
        34.01 / 0.938 & 35.71 / 0.956 & 38.09 / 0.951 & 41.26 / 0.981 \\
        \bottomrule
    \end{tabular}
    }
    \caption{\textbf{Quantitative comparison} of denoising performance. PSNR/SSIM ($\uparrow$) were used.} 
    \label{reb:tab:denoising_other_datasets}
\end{table}
\label{sec:ablation}
\begin{figure*}
    \centering
    \begin{subfigure}[]{0.44\textwidth}
        \centering
        \includegraphics[width=\textwidth]{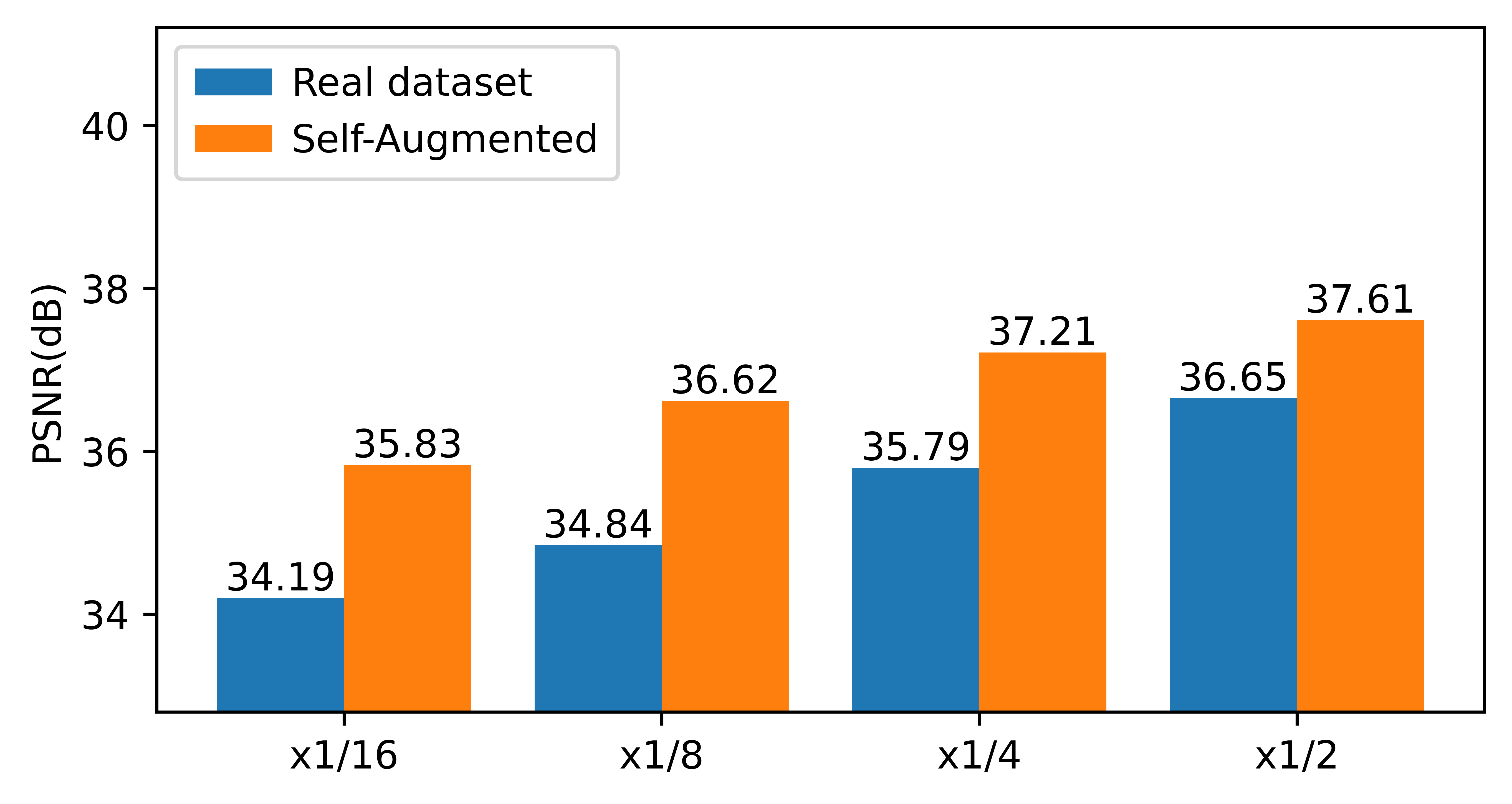}
        \caption{PSNR comparison across dataset sizes}
        \label{fig:self_aug_datasize}
    \end{subfigure}
    \centering
    \begin{subfigure}[]{0.48\textwidth}
        \centering
        \includegraphics[width=\textwidth]{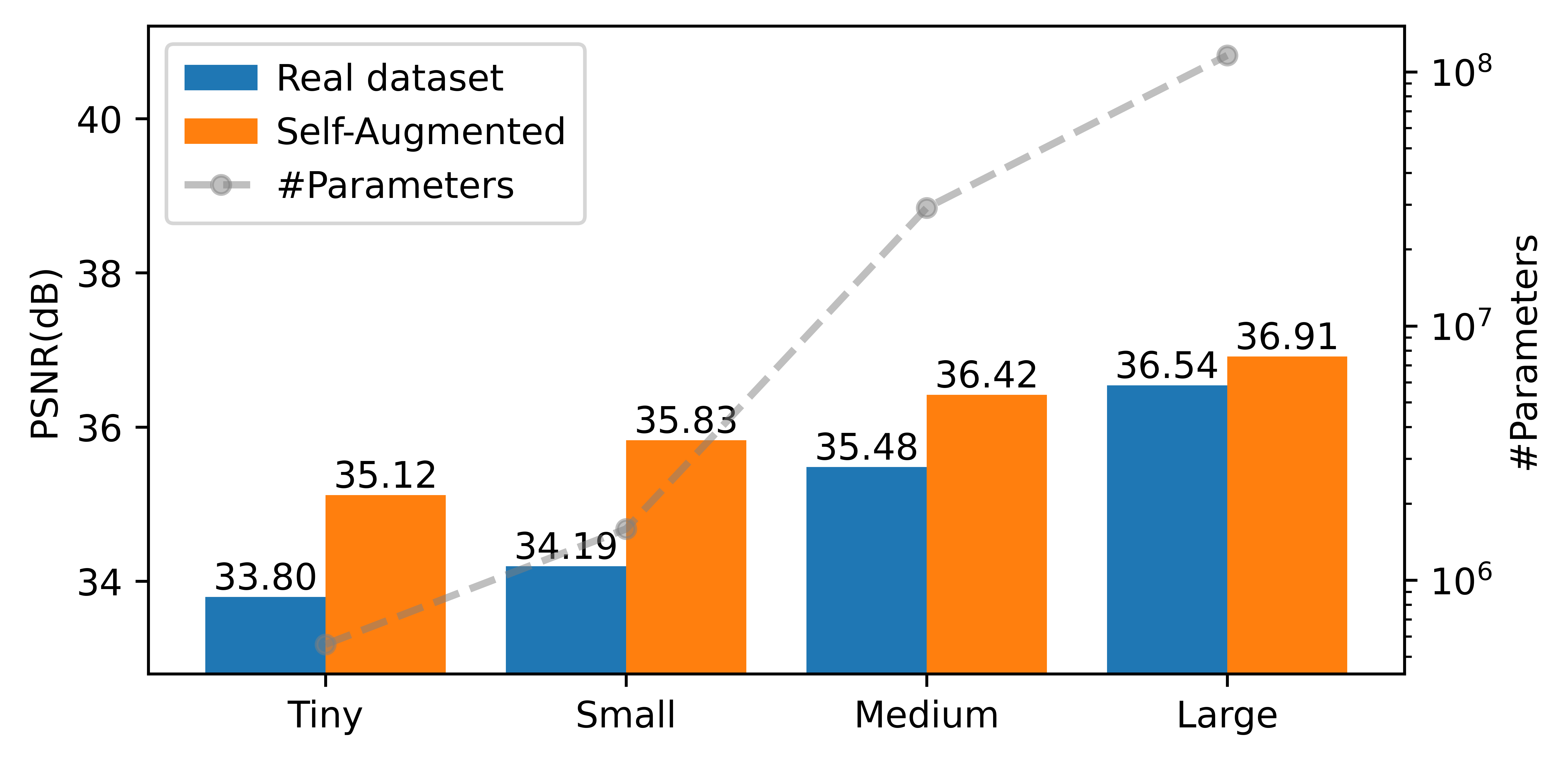}
        \caption{PSNR comparison across model sizes}
        \label{fig:self_aug_modelsize}
    \end{subfigure}
    \caption{
    \textbf{Denoising performance trends by model and dataset size.} Figure~\ref{fig:self_aug_datasize} presents a performance comparison across dataset sizes for NAFNet-Small. Figure~\ref{fig:self_aug_modelsize} presents a performance comparison across model sizes on the ×1/16 dataset.
    }
    \label{fig:results_self_augmented}
\end{figure*}

\paragraph{Ablation in Image denoising.} As demonstrated above, the components (guidance, refine loss) in {GuidNoise} contribute to synthesized noise quality, even when using guidance that differs from the input clean image. This flexibility allow to augment limited real noise datasets by squaring them, which we call \textbf{Self-augmentation.}
To demonstrate {GuidNoise}'s efficacy in image denoising, we train NAFNet of various sizes on datasets of different scales, with and without self-augmentation. For self-augmented training, each batch contains an equal mix of real and synthetic data.

We evaluate denoising performance using peak PSNR across various size of training datasets ($\times1/2$, $\times1/4$, $\times1/8$, $\times1/16$ of the original dataset) and model sizes (Tiny, Small, Medium, Large). 
For each case, we investigate the improvement of denoising performance via self-augmentation from the original dataset. 
Detailed results are presented with visual trends shown in Figure~\ref{fig:results_self_augmented}.
Self-augmentation improves denoising performance compared to the original real data across various scenarios. It is particularly effective for small models or small datasets in scenarios close to real-world conditions. For NAFNet-small, a self-augmented model trained on 1/8 of real data achieves a PSNR of 36.62dB, nearly equivalent to the 36.65dB PSNR of a model trained on 1/2 of real data. Moreover, with only 1/16 of the real dataset, a self-augmented NAFNet-Small (1.59M parameters) outperforms NAFNet-Medium (29.16M parameters) trained solely on real data, achieving 35.83dB PSNR compared to 35.48dB PSNR, respectively. 
\begin{table}[]
    \centering
    \aboverulesep=0ex
    \belowrulesep=0ex
    \resizebox{0.8\linewidth}{!}{
    \begin{tabular}{l | cc }
    \toprule 
    Method &  KLD\text{$\downarrow$} & AKLD\text{$\downarrow$} \\ \midrule
    Baseline $({\mathbf{x}}, \mathbf{c})$, $\mathcal{L}_\text{Diffusion}$ & 0.080 & 0.150 \\
    + Refine loss $\mathcal{L}_\text{Refine}$ & 0.028 & 0.163 \\
    \midrule
    + \textbf{Guidance} $(\mathbf{x}_\text{r}, \mathbf{c}_\text{r})$ & 0.050 & 0.118 \\
    + \textbf{Refine loss} $\mathcal{L}_\text{Refine}$ & \textbf{0.014} & \textbf{0.113}  \\
    \bottomrule
    \end{tabular}
    }
    \caption{\textbf{Ablation studies.} Evaluating the effectiveness of guidance and refine loss on SIDD-Validation.}
    \label{tab:diffusion_ablation}
\end{table}

Regarding the generalization for denoising, in Table~\ref{reb:tab:denoising_other_datasets}, we provide a comparison between real and synthetic settings for PolyU and Nam datasets.

The self-augmentation experiments verify the effectiveness (see Figure~\ref{fig:results_self_augmented}), where the synthesized noisy images in the unpaired scenarios improve the performance.
This demonstrates that self-augmentation via {GuidNoise} can relax the cost of real noisy datasets or maintain the denoising performance even with few parameters.

\paragraph{Noisy Image Synthesis and Refine Loss.} 
To see the effectiveness of the proposed noise guidance and \textbf{refine loss}, we demonstrate the ablation in Table~\ref{tab:diffusion_ablation}. 
The guidance image reduces KLD from 0.080 to 0.050 and AKLD from 0.150 to 0.118 compared to the baseline which uses only input clean images. The refine loss further reduces KLD from 0.050 to 0.014. These results demonstrate that our cascade decoding architecture and noise-aware guidance significantly improve noise synthesis.
Furthermore, a recent study PUCA~\cite{jang2023puca} trained in SIDD-Medium achieved an unsupervised performance of PSNR 37.49, slightly higher than our self-augmentation($\times 1/8$) PSNR 37.48. While PUCA uses 320 images, our approach employs a smaller subset ($\times 1/8$) from SIDD-Validation, not SIDD-Medium, so that it consists of only five pairs.
\section{Conclusion}
This paper proposes {GuidNoise}, a single-pair guided diffusion method for enhanced and generalized noise synthesis in image denoising. We mitigate the existing methods' prerequisites by using just a single guidance image pair, easily obtainable from the training set. Our approach introduces a new guidance-aware affine feature modification and noise-aware refine loss that leverages the inherent potential of diffusion models for noisy image generation, allowing the backward process to generate more realistic noise distributions. Additionally, our cascade decoding architecture focuses on the noise distribution of the guidance, thus improving generalization performance across various guidance images. Through these innovations, {GuidNoise} demonstrates notable noise similarity on various datasets. 

\section{Acknowledgements}
This work was supported by the Institute of Information $\&$ Communications Technology Planning $\&$ Evaluation (IITP) grant funded by the Korea government (MSIT) [RS-2021-II211341, Artificial Intelligence Graduate School Program (Chung-Ang University) and RS-2022-II220124, Development of Artificial Intelligence Technology for Self-Improving Competency-Aware Learning Capabilities]. SNUAILAB, corp, also supports this work.

\bibliography{aaai2026}
\clearpage
\setcounter{page}{1}
\setcounter{section}{0}
\setcounter{figure}{0}
\setcounter{table}{0}
\setcounter{equation}{0}
\begin{figure}[b!]
    \centering
    \includegraphics[width=0.48\textwidth]{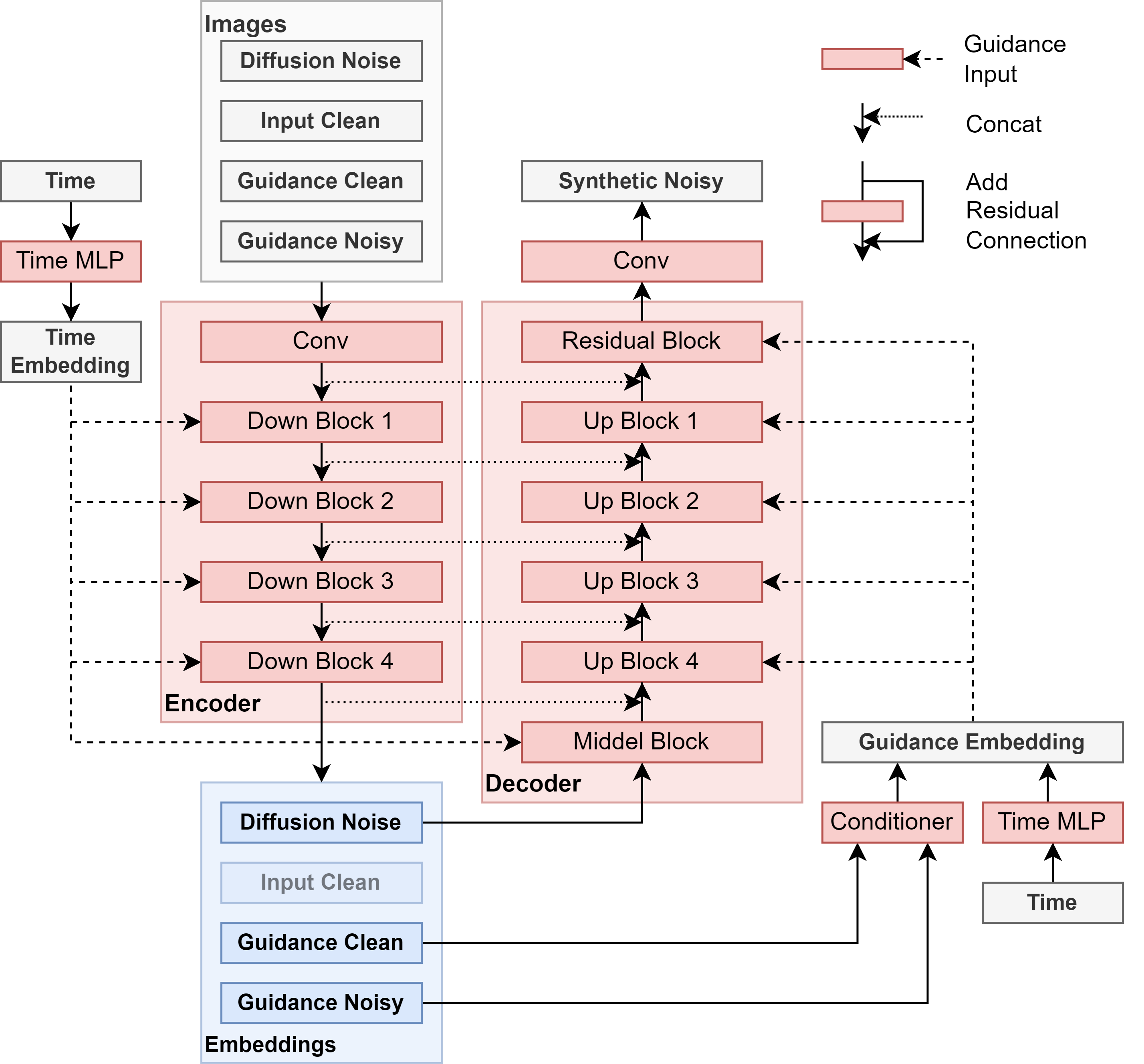}
    \caption{Architecture overview of GuidNoise.}
    \label{fig:supple_model_1}
\end{figure}
\begin{figure}[b!]
    \centering
    \includegraphics[width=0.48\textwidth]{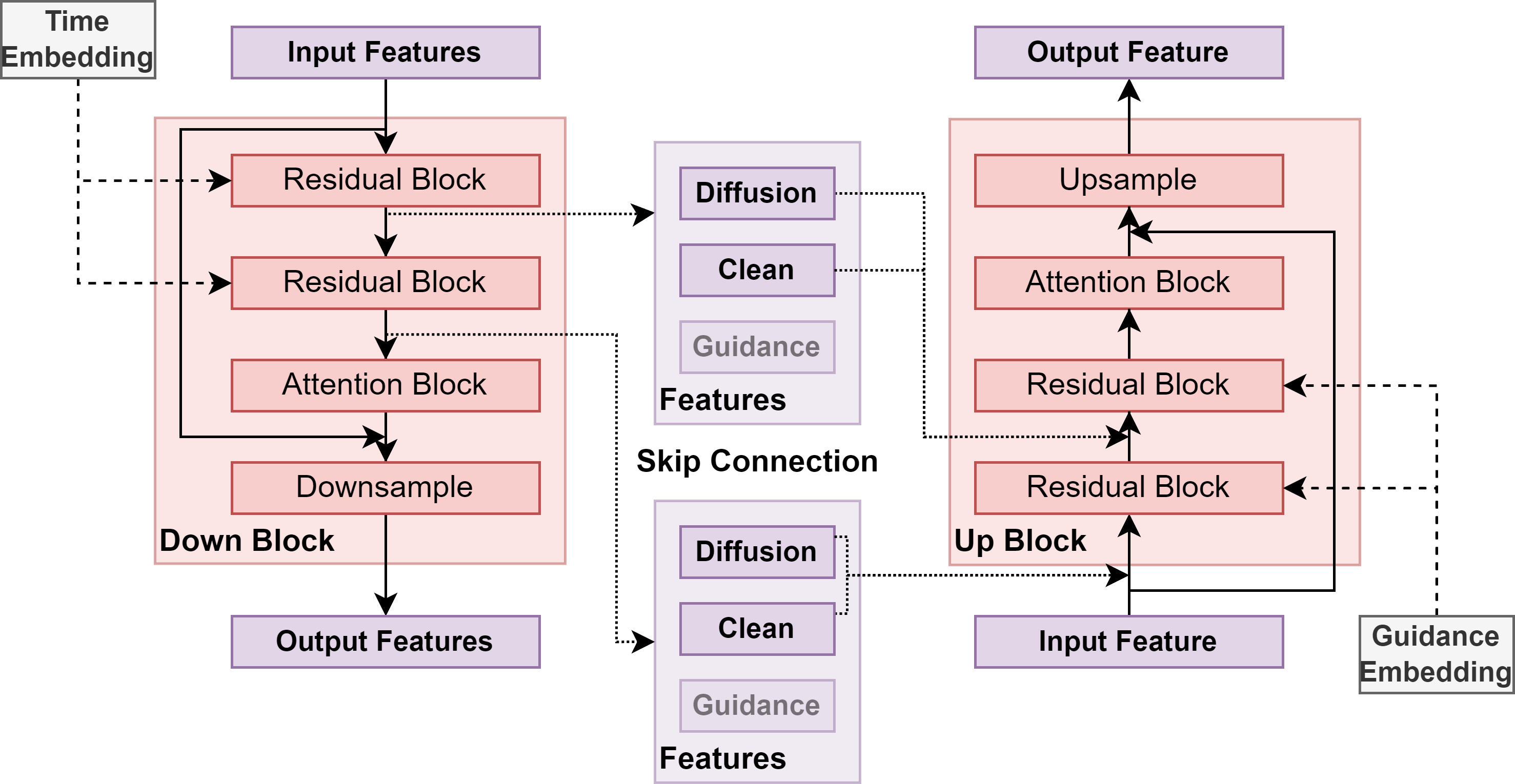}
    \caption{Architecture of Down/Up Blocks with skip connection.}
    \label{fig:supple_model_2}
\end{figure}
\begin{figure}[b!]
    \centering
    \includegraphics[width=0.45\textwidth]{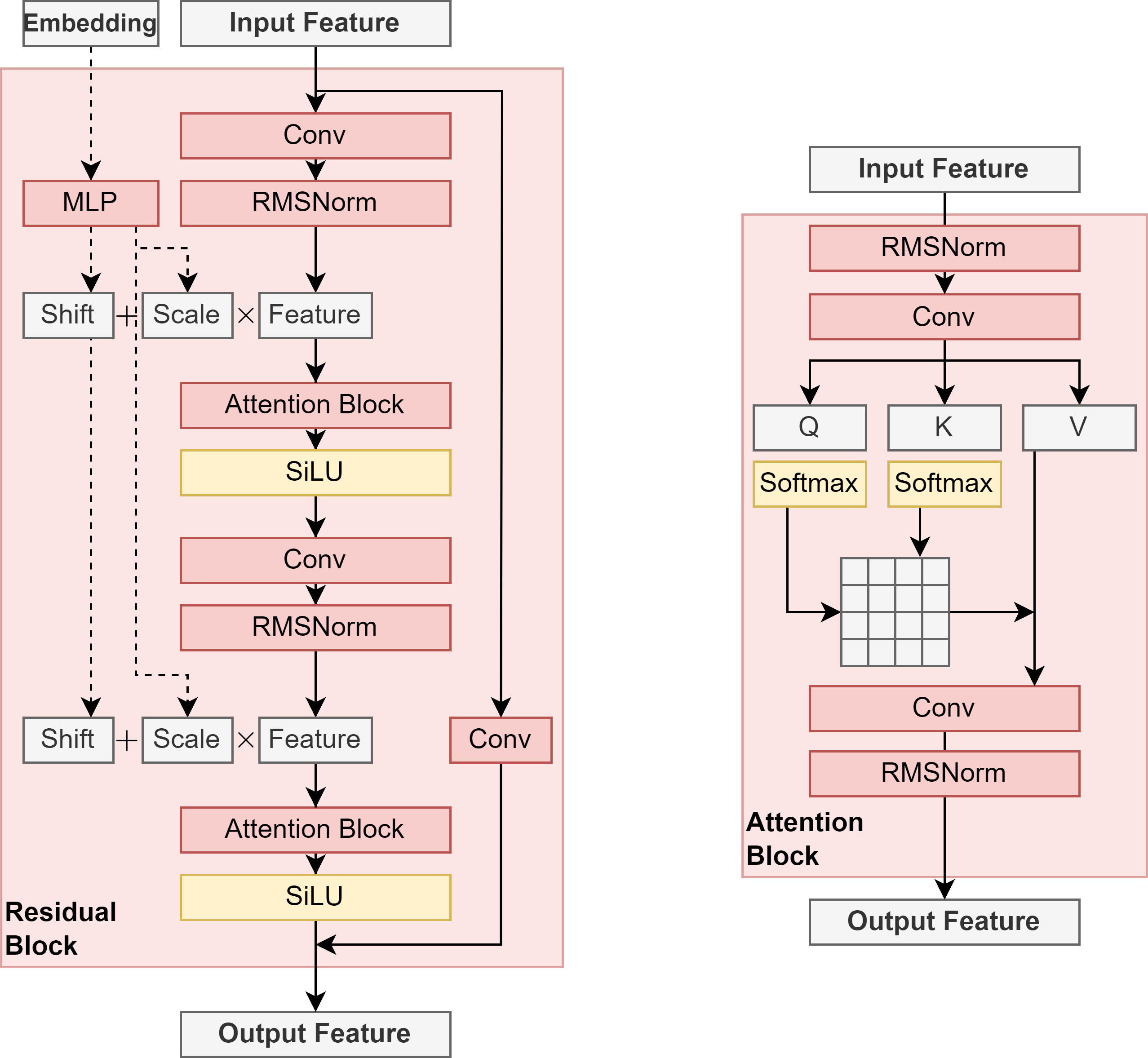}
    \caption{Architecture of Residual/Attention Blocks.}
    \label{fig:supple_model_5}
\end{figure}
\section{Model architecture and size}
\label{sec:model_architecture}
The proposed reference-guided noise synthesis diffusion (GuidNoise) is based on a conditional U-Net architecture from DDPM~\cite{DDPM_Diffusion:ho2020denoising}. The detailed architecture of our model is shown in Figure~\ref{fig:supple_model_1}, Figure~\ref{fig:supple_model_2}, and Figure~\ref{fig:supple_model_5}. Specifically, the encoding process in Eq.~(\ref{eq:encoder}) of the main manuscript corresponds to the Encoder in Figure~\ref{fig:supple_model_1}. The guidance-aware affine feature modification (GAFM) $\tau$ in Eq.~(\ref{eq:guidance_module}) corresponds to the Conditioner in Figure~\ref{fig:supple_model_2}. The cascade decoding process in Eq.~(\ref{eq:decoder}) corresponds to the decoding process in Figure~\ref{fig:supple_model_5}. The number of parameters for GuidNoise and others can be compared as shown in Table~\ref{reb:tab:params}, where we note that lightweight models such as C2N, Flow-sRGB, and NAFlow have performance limitations, while NeCA includes models with varying parameter sizes.
\vspace{3em}
 
\section{Self-augmentation}
\label{sec:supple:self-augmentation}
\noindent\textbf{Model Variants} GuidNoise allows us to augment limited real noise datasets, which is referred to as Self-augmentation. We conduct the experiments for self-augmented training across various dataset size and model size of NAFNet~\cite{NAFNet:chen2022simple}. The specific model configuration of each size in Table~\ref{tab:nafnet_configurations}.

\noindent\textbf{Results} Due to the varying results depending on model size and data size, we select representative trend results as shown in  Figure~\ref{fig:results_self_augmented} in the main manuscript. 
We present all results, including the remaining trends on Figure~\ref{fig:supple_results_self_augmented}. 
Self-augmentation via GuidNoise shows significant benefits, especially with limited training data and smaller model architectures.
\vspace{3em}
\begin{table}[!b]
\centering
\resizebox{0.47\textwidth}{!}{
\begin{tabular}{l|c|c|c|c}
\hline
\multirow{3}{*}{\textbf{\shortstack{Model\\Type}}} & 
\multirow{3}{*}{\textbf{\shortstack{Channel\\Width}}} &
\multirow{3}{*}{\textbf{\shortstack{\#Encoder\\Blocks}}} &
\multirow{3}{*}{\textbf{\shortstack{\#Middle\\Blocks}}} &
\multirow{3}{*}{\textbf{\shortstack{\#Decoder\\Blocks}}} \\
 & & & & \\ & & & & \\ \hline
Tiny   & 8  & 1, 1, 1, 2 & 3  & 1, 1, 1, 1 \\ \hline
Small  & 12 & 1, 1, 2, 3 & 4  & 1, 1, 1, 1 \\ \hline
Medium & 32 & 2, 2, 4, 8 & 12 & 2, 2, 2, 2 \\ \hline
Large  & 64 & 2, 2, 4, 8 & 12 & 2, 2, 2, 2 \\ \hline
\end{tabular}
}
\caption{Configurations of NAFNet variants.}
\label{tab:nafnet_configurations}
\end{table}
\begin{table}[!b]
\centering
\resizebox{0.45\textwidth}{!}{
\begin{tabular}{lcccccc}
\hline
\textbf{Model}      & C2N & Flow-sRGB & NAFlow\\ \hline
\textbf{\#Params}   & 2.2M & 6K & 1.1M \\ \hline \hline
\textbf{Model}      & NeCA-S & NeCA-W & Ours \\ \hline
\textbf{\#Params}   & $5 \times (7.8\text{M}^c)$ & $5 \times (7.8\text{M}^c + \alpha)$ &  62M\\ \hline
\end{tabular}
}
\caption{Comparison of model parameters. The remaining parameter size of NeCA is set to $\alpha=263\text{K}^a + 42\text{K}^b$.  }
\label{reb:tab:params}
\end{table}

\def\tmpheight{4cm}
\begin{figure*}
    \centering
    \begin{subfigure}[]{0.49\textwidth}
        \centering
        \includegraphics[height=\tmpheight]{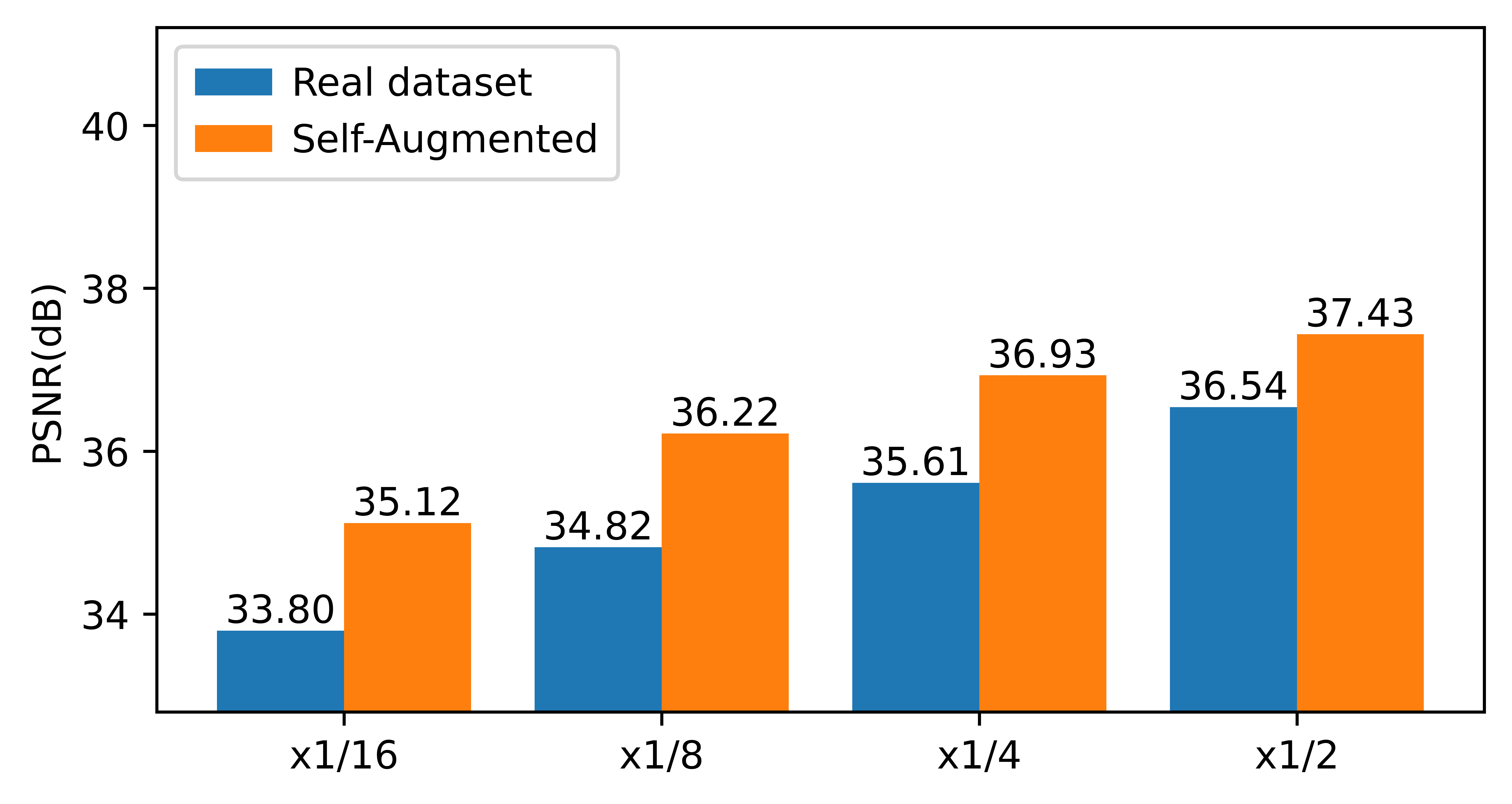}
        \caption{PSNR comparison across dataset sizes on NAFNet-Tiny}
        \label{fig:supple_self_aug_tiny}
    \end{subfigure}
    \centering
    \begin{subfigure}[]{0.49\textwidth}
        \centering
        \includegraphics[height=\tmpheight]{imgs/self-aug/PSNR_comparison_x1_16.png}
        \caption{PSNR comparison across model sizes on $\times1/16$ data}
        \label{fig:supple_self_aug_16}
    \end{subfigure}
    \\
    \centering
    \begin{subfigure}[]{0.49\textwidth}
        \centering
        \includegraphics[height=\tmpheight]{imgs/self-aug/PSNR_comparison_small.png}
        \caption{PSNR comparison across dataset sizes on NAFNet-Small}
        \label{fig:supple_self_aug_small}
    \end{subfigure}
    \centering
    \begin{subfigure}[]{0.49\textwidth}
        \centering
        \includegraphics[height=\tmpheight]{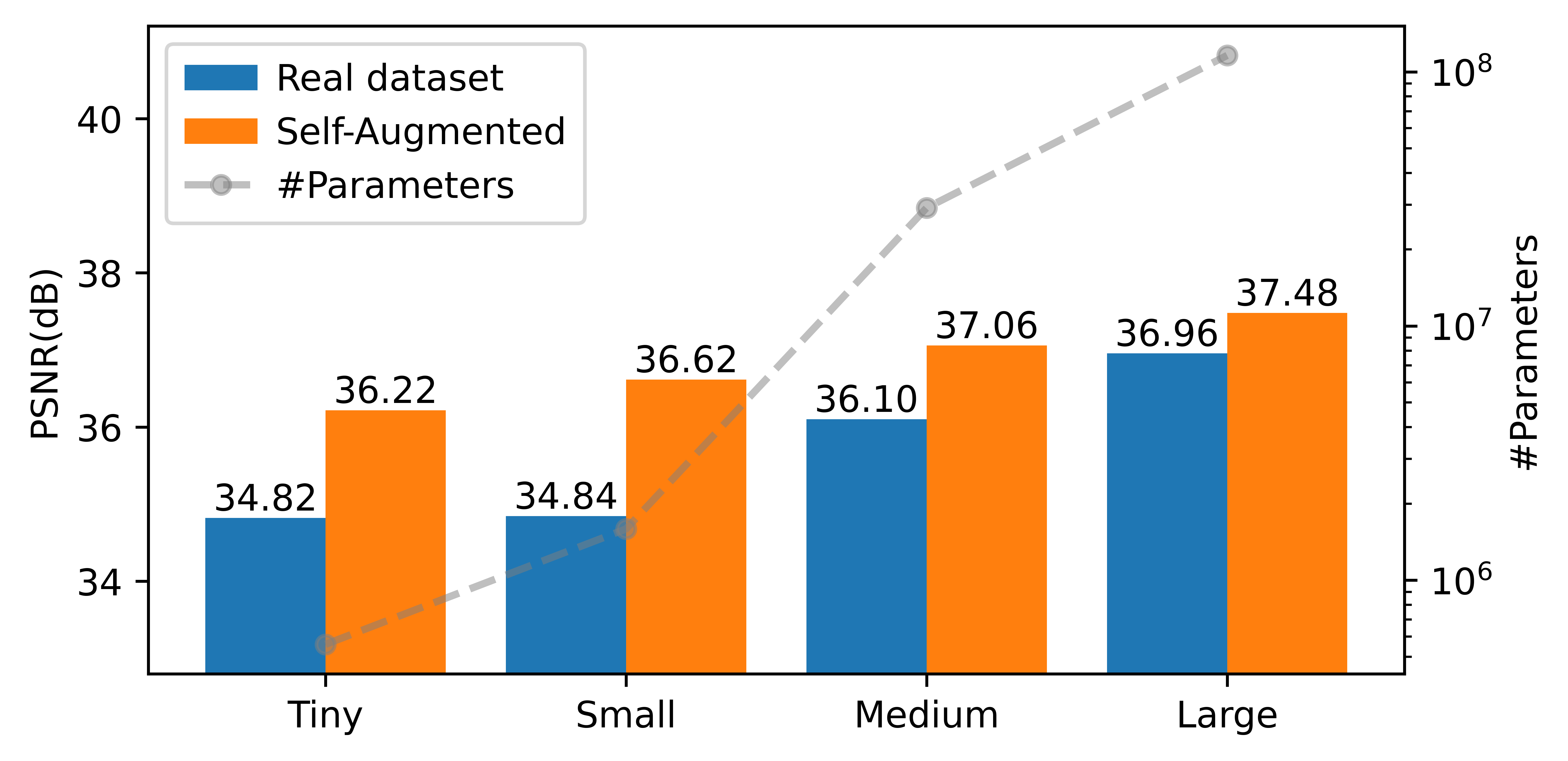}
        \caption{PSNR comparison across model sizes on $\times1/8$ data}
        \label{fig:supple_self_aug_8}
    \end{subfigure}
    \\
    \centering
    \begin{subfigure}[]{0.49\textwidth}
        \centering
        \includegraphics[height=\tmpheight]{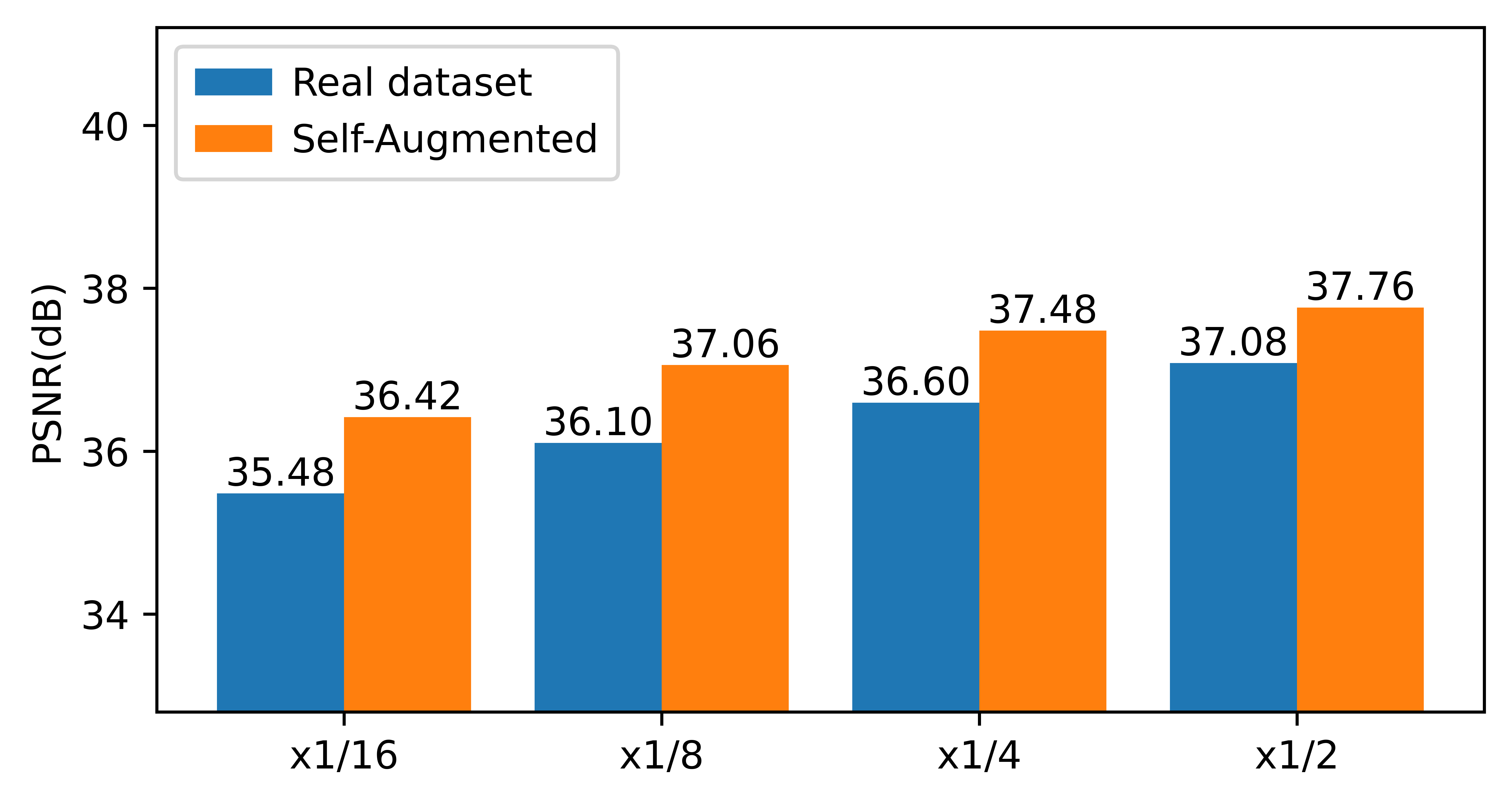}
        \caption{PSNR comparison across dataset sizes on NAFNet-Medium}
        \label{fig:supple_self_aug_medium}
    \end{subfigure}
    \centering
    \begin{subfigure}[]{0.49\textwidth}
        \centering
        \includegraphics[height=\tmpheight]{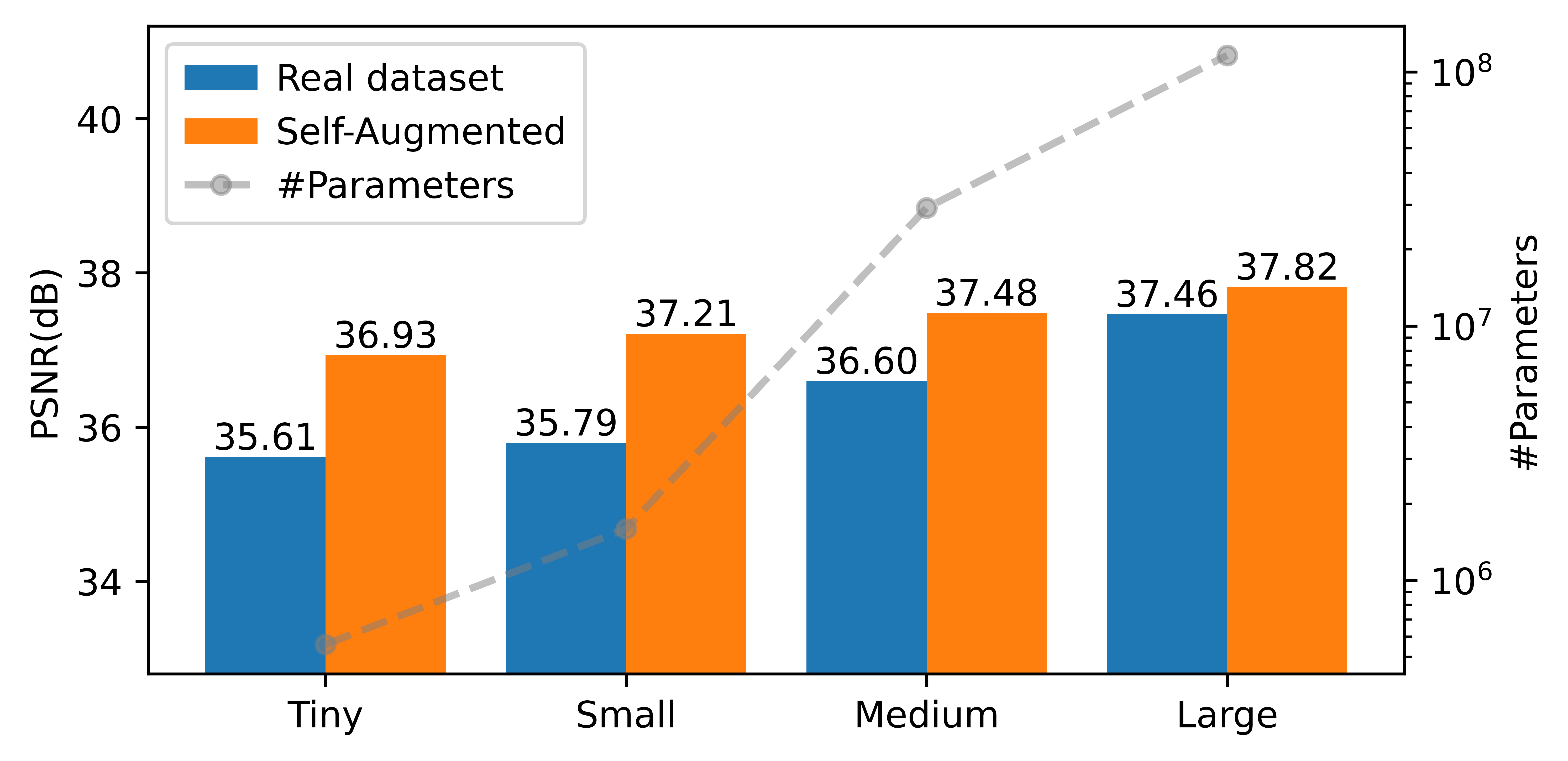}
        \caption{PSNR comparison across model sizes on $\times1/4$ data}
        \label{fig:supple_self_aug_4}
    \end{subfigure}
    \\
    \centering
    \begin{subfigure}[]{0.49\textwidth}
        \centering
        \includegraphics[height=\tmpheight]{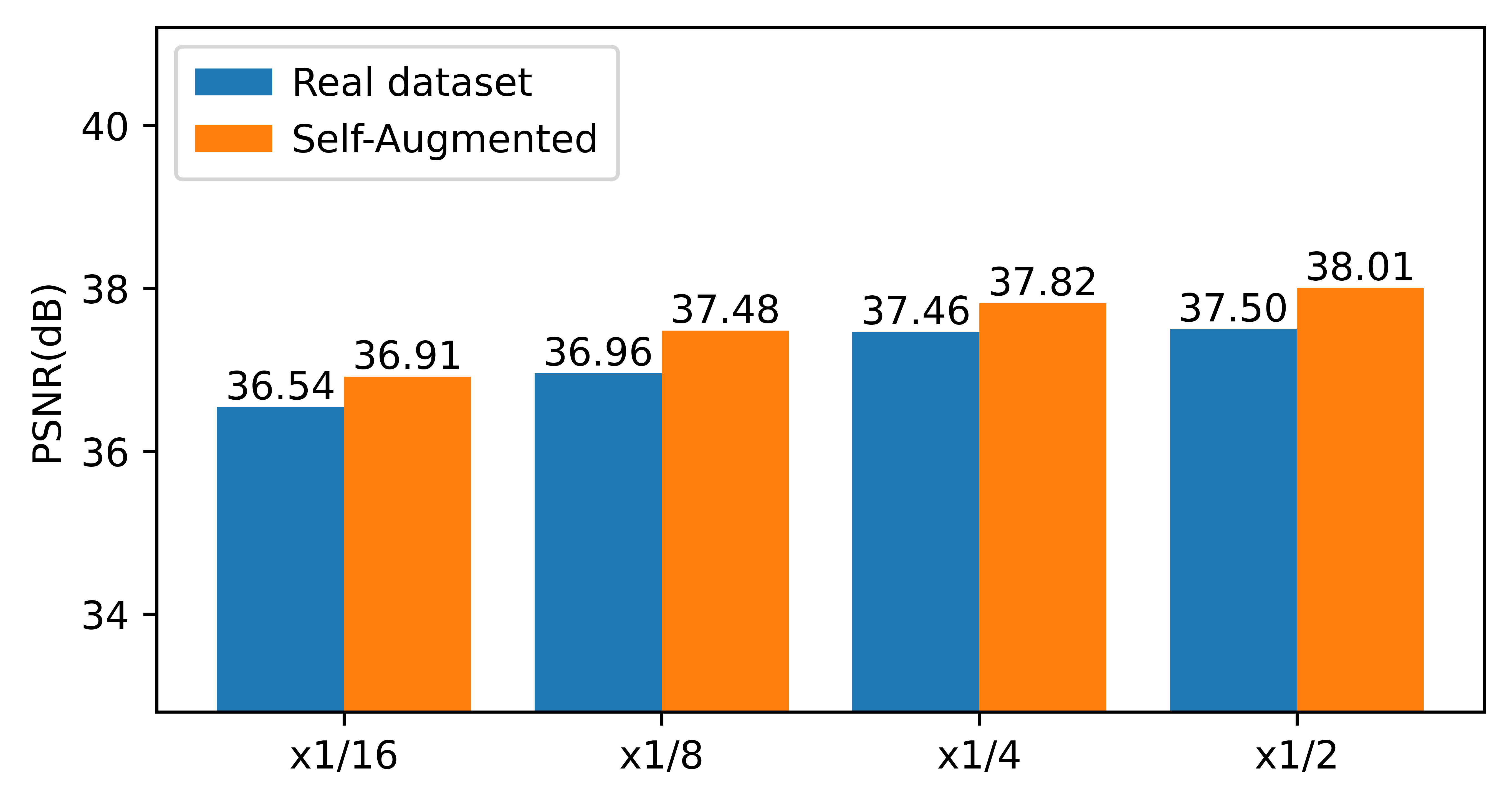}
        \caption{PSNR comparison across dataset sizes on NAFNet-Large}
        \label{fig:supple_self_aug_large}
    \end{subfigure}
    \centering
    \begin{subfigure}[]{0.49\textwidth}
        \centering
        \includegraphics[height=\tmpheight]{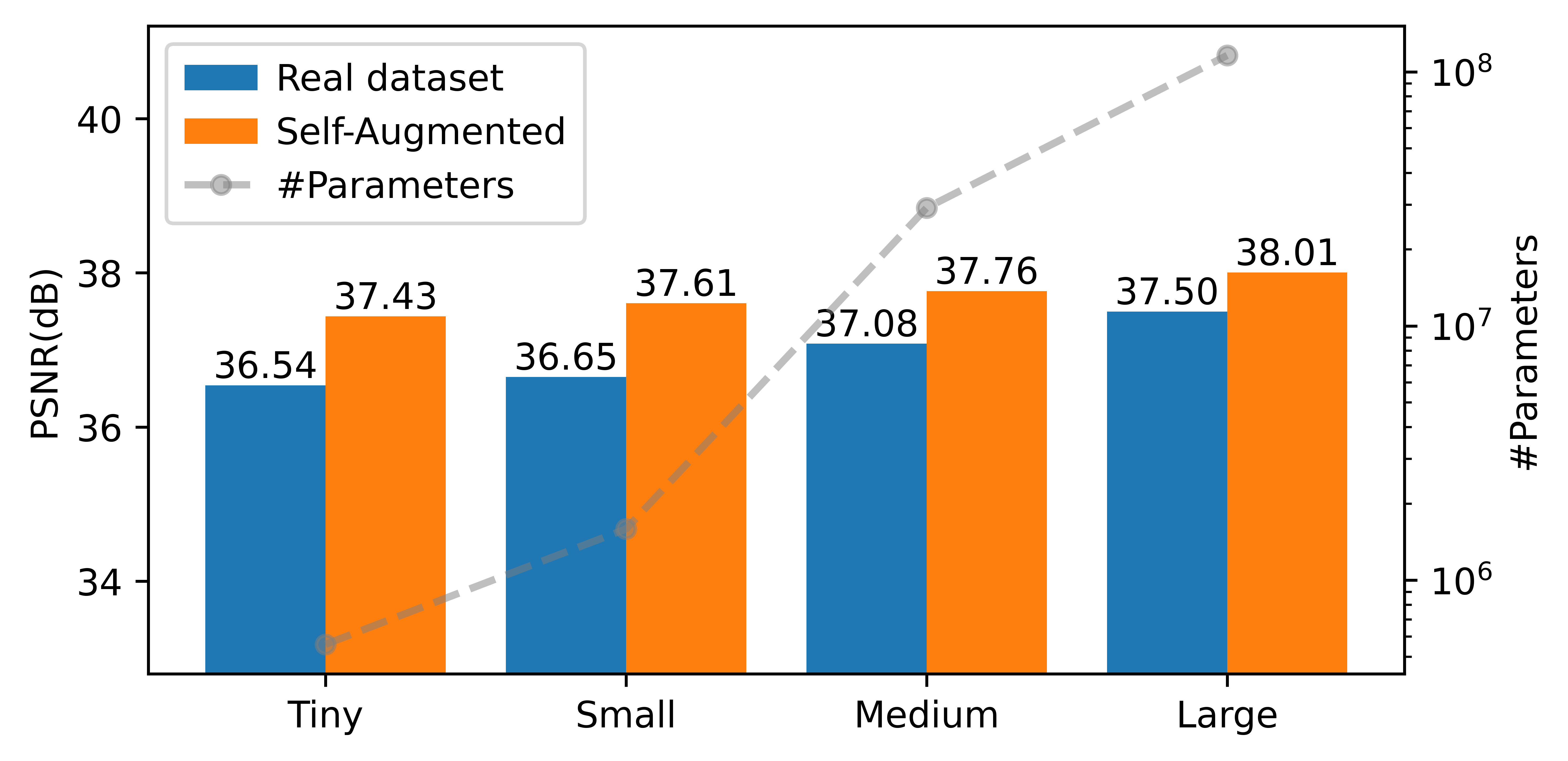}
        \caption{PSNR comparison across model sizes on $\times1/2$ data}
        \label{fig:supple_self_aug_2}
    \end{subfigure}
    \caption{
    \textbf{Denoising performance trends} across model size (left) and dataset size (right). The smaller the training data or the smaller the model, the larger the self-augmentation effect tends to be.}
    \label{fig:supple_results_self_augmented}
\end{figure*}

\begin{table*}[]
    \centering
    \resizebox{1\textwidth}{!}{
    \begin{tabular}{c | c c | c c | c c | c c }
    \toprule 
    \multirow{4}{*}{\shortstack{Dataset \\ Size}} &  \multicolumn{2}{c|}{NAFNet-Tiny} & \multicolumn{2}{c|}{NAFNet-Small} & \multicolumn{2}{c|}{NAFNet-Medium}  & \multicolumn{2}{c}{NAFNet-Large} \\
    \cmidrule{2-9} 
    & Real dataset & Self-augmented 
    & Real dataset & Self-augmented 
    & Real dataset & Self-augmented 
    & Real dataset & Self-augmented \\ 
    \cmidrule{2-9}
    & PSNR\text{$\uparrow$} & PSNR\text{$\uparrow$} 
    & PSNR\text{$\uparrow$} & PSNR\text{$\uparrow$} 
    & PSNR\text{$\uparrow$} & PSNR\text{$\uparrow$} 
    & PSNR\text{$\uparrow$} & PSNR\text{$\uparrow$} 
    \\ 
    \midrule
    $\times$1/16 & 33.80 & 35.12 & 34.19 & 35.83 & 35.48 & 36.42 & 36.54 & 36.91 \\
    $\times$1/8 & 34.82 & 36.22 & 34.84 & 36.62 & 36.10 & 37.06 & 36.96 & 37.48 \\
    $\times$1/4   & 35.61 & 36.93 & 35.79 & 37.21 & 36.60 & 37.48 & 37.46 & 37.82 \\
    $\times$1/2   & 36.54 & 37.43 & 36.65 & 37.61 & 37.08 & 37.76 & 37.50 & 38.01 \\
    \midrule
    \#Parameters & \multicolumn{2}{c|}{0.56M} & \multicolumn{2}{c|}{1.59M} & \multicolumn{2}{c|}{29.16M} & \multicolumn{2}{c}{115.98M} \\
    \bottomrule
    \end{tabular}
    }
    \caption{
    \textbf{Denoising performance comparison across model and dataset sizes.} PSNR values on SIDD-Validation for different model configurations (Tiny, Small, Medium, Large) and dataset sizes (×1/16, ×1/8, ×1/4, ×1/2) trained on real dataset and self-augmented datasets.
    }
    \vspace{-2mm}
    \label{tab:denoiser_self_augmented_performance}
\end{table*}
\begin{figure*}[t]
    \centering
    \begin{subfigure}[b]{0.58\textwidth}
        \begin{subfigure}[b]{\textwidth}
            \centering
            \includegraphics[width=\textwidth]{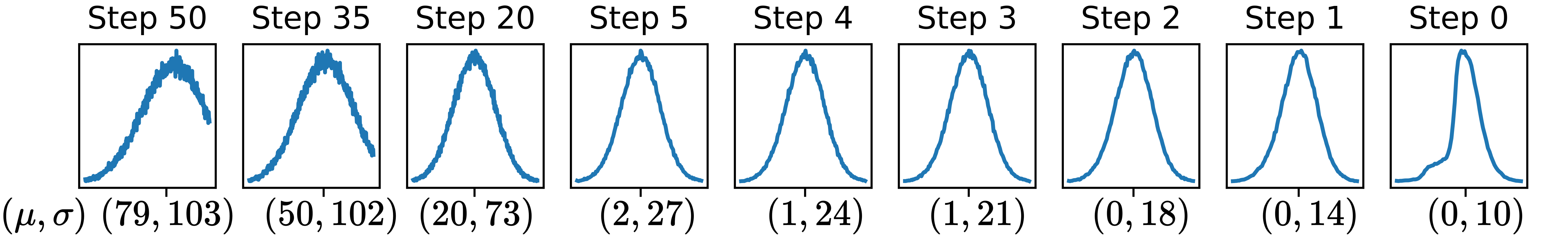}
            \caption{wo/ Refine Loss}
            \label{fig:histogram_wo_refine_loss}
        \end{subfigure}
        \\
        \centering
        \begin{subfigure}[b]{\textwidth}
            \centering
            \includegraphics[width=\textwidth]{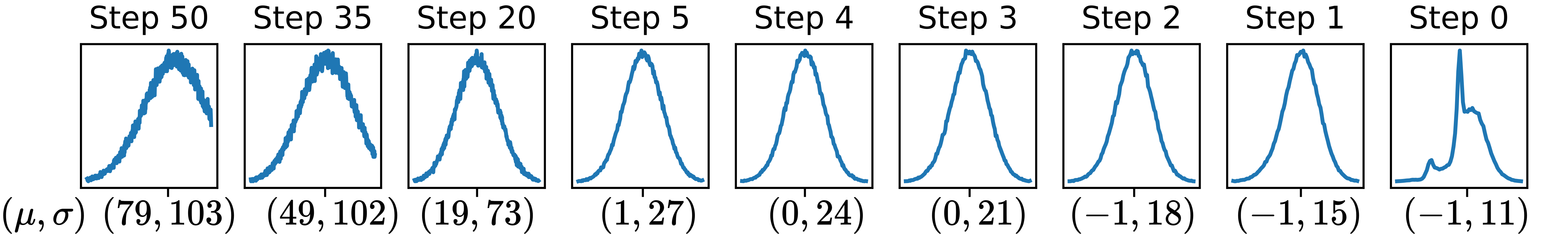}
            \caption{w/ Refine Loss}
            \label{fig:histogram_w_refine_loss}
        \end{subfigure}
    \end{subfigure}
    \centering
    \begin{subfigure}[b]{0.26\textwidth}
        \centering
        \includegraphics[width=\textwidth]{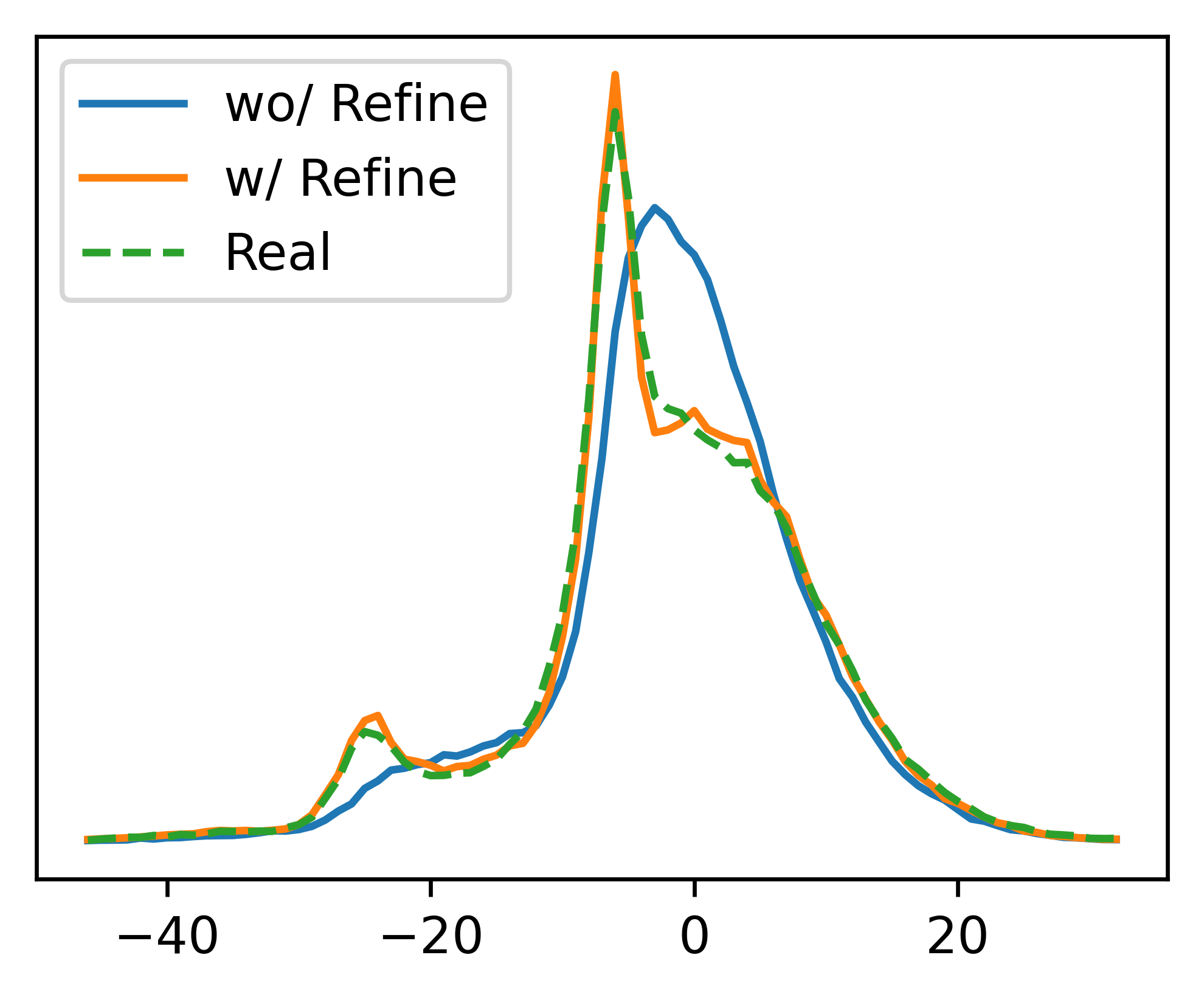}
        \caption{Noise distribution comparison}
        \label{fig:histogram_comparision}
    \end{subfigure}
    \caption{
    \textbf{Noise Distribution Visualization.} Figures~\ref{fig:histogram_wo_refine_loss} and ~\ref{fig:histogram_w_refine_loss} show the noise distribution changes per time step. Figure~\ref{fig:histogram_comparision} contrasts noise distributions of real noisy data, images generated with refine loss (w/ Refine), and without (wo/ Refine).
    }
    \label{fig:histogram}
\end{figure*}
\def\imgsize{0.155} 
\begin{figure*}[h!]
    \centering
    \begin{subfigure}[b]{0.03\textwidth}
        \centering
        \caption*{\raisebox{1.2\height}{\rotatebox{90}{\scriptsize }}}
    \end{subfigure}
    \begin{subfigure}[b]{\imgsize\textwidth}
        \centering
        \includegraphics[width=\textwidth]{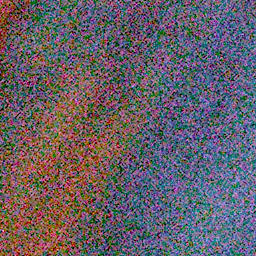}
    \end{subfigure}
    \begin{subfigure}[b]{\imgsize\textwidth}
        \centering
        \includegraphics[width=\textwidth]{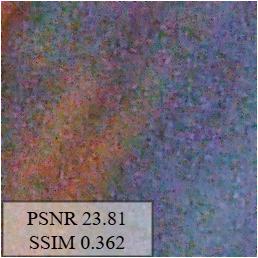}
    \end{subfigure}
    \begin{subfigure}[b]{\imgsize\textwidth}
        \centering
        \includegraphics[width=\textwidth]{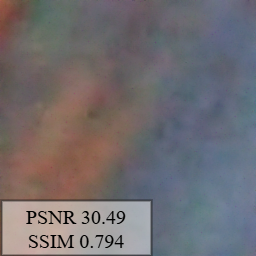}
    \end{subfigure}
    \begin{subfigure}[b]{\imgsize\textwidth}
        \centering
        \includegraphics[width=\textwidth]{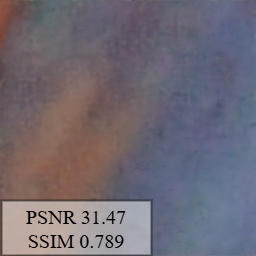}
    \end{subfigure}
    \begin{subfigure}[b]{\imgsize\textwidth}
        \centering
        \includegraphics[width=\textwidth]{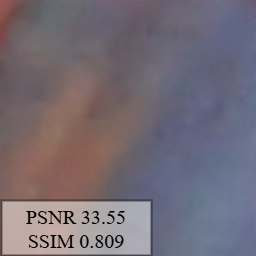}
    \end{subfigure}
    \begin{subfigure}[b]{\imgsize\textwidth}
        \centering
        \includegraphics[width=\textwidth]{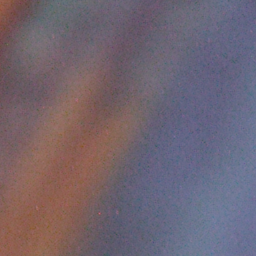}
    \end{subfigure}
    \\
    \begin{subfigure}[b]{0.03\textwidth}
        \centering
        \caption*{\raisebox{1.0\height}{\rotatebox{90}{\scriptsize }}}
    \end{subfigure}
    \begin{subfigure}[b]{\imgsize\textwidth}
        \centering
        \includegraphics[width=\textwidth]{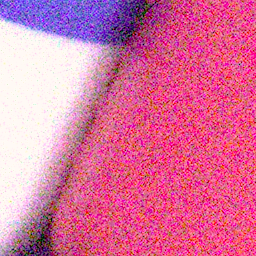}
    \end{subfigure}
    \begin{subfigure}[b]{\imgsize\textwidth}
        \centering
        \includegraphics[width=\textwidth]{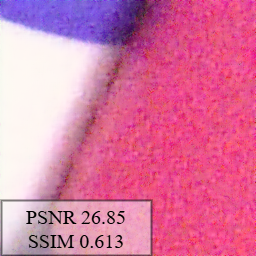}
    \end{subfigure}
    \begin{subfigure}[b]{\imgsize\textwidth}
        \centering
        \includegraphics[width=\textwidth]{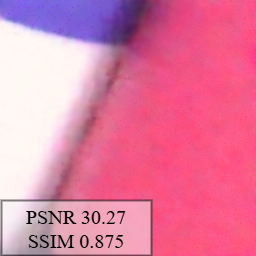}
    \end{subfigure}
    \begin{subfigure}[b]{\imgsize\textwidth}
        \centering
        \includegraphics[width=\textwidth]{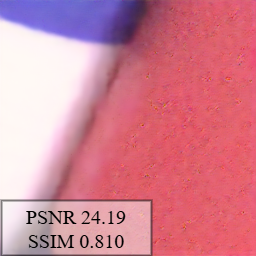}
    \end{subfigure}
    \begin{subfigure}[b]{\imgsize\textwidth}
        \centering
        \includegraphics[width=\textwidth]{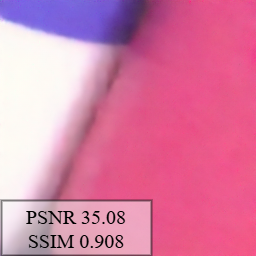}
    \end{subfigure}
    \begin{subfigure}[b]{\imgsize\textwidth}
        \centering
        \includegraphics[width=\textwidth]{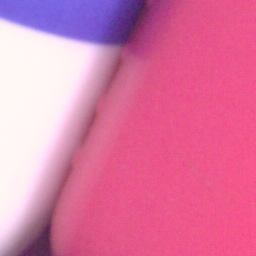}
    \end{subfigure}
    \begin{subfigure}[b]{0.03\textwidth}
        \centering
        \caption*{}
    \end{subfigure}
    \begin{subfigure}[b]{\imgsize\textwidth}
        \centering
        \caption*{\scriptsize Noisy}
    \end{subfigure}
    \begin{subfigure}[b]{\imgsize\textwidth}
        \centering
        \caption*{\scriptsize C2N}
    \end{subfigure}
    \begin{subfigure}[b]{\imgsize\textwidth}
        \centering
        \caption*{\scriptsize NeCA-W}
    \end{subfigure}
    \begin{subfigure}[b]{\imgsize\textwidth}
        \centering
        \caption*{\scriptsize NAFlow}
    \end{subfigure}
    \begin{subfigure}[b]{\imgsize\textwidth}
        \centering
        \caption*{\scriptsize Ours}
    \end{subfigure}
    \begin{subfigure}[b]{\imgsize\textwidth}
        \centering
        \caption*{\scriptsize Clean}
    \end{subfigure}
    \\
    \caption{\textbf{Qualitative comparison of denoising results} on the SIDD-Validation dataset.}
    \label{fig:denoiser_performance}
\end{figure*}

\section{Refine Loss}
\label{sec:supple:refine_loss}
We validate the proposed refine loss that refines the diffusion process to consider the noise distribution of synthesized images. As shown in Figure~\ref{fig:histogram}, the synthesized noise distribution at Step 0 with the refine loss looks similar to the real noise distribution, while that without the refine loss can be seen as very different from the real noise  distribution. 
During DDIM sampling process, the noise distribution primarily undergoes large-scale adjustments in most time steps, mainly in the form of shifts and scaling. However, fine adjustments occur within the final 1–2 steps. Leveraging this insight, we define the refine loss and set the differentiable sampling step $T_\text{split}$ to last two sampling steps. 

\section{Qualitative denoising results}
We present the visualization of denoised noisy images in Figure~\ref{fig:denoiser_performance}. Notably, the denoised images from the denoising model trained with synthesized images from GuidNoise are visually closest to the clean image. This indirectly supports the realistic synthesis performance of GuidNoise.
\end{document}